\pdfoutput=1
\documentclass[11pt]{article}

\usepackage{EMNLP2023}

\usepackage{times}
\usepackage{latexsym}
\usepackage{placeins}

\usepackage[T1]{fontenc}
\usepackage[utf8]{inputenc}

\usepackage{microtype}

\usepackage{inconsolata}

\usepackage{amsmath}
\usepackage{amsfonts}
\usepackage{hhline}
\usepackage{multirow}
\usepackage{makecell}
\usepackage{float}

\usepackage[T1]{fontenc}    % use 8-bit T1 fonts
\usepackage{url}            % simple URL typesetting
\usepackage{booktabs, multirow, tabularx}       % professional-quality tables
\usepackage{amsfonts}       % blackboard math symbols
\usepackage{nicefrac}       % compact symbols for 1/2, etc.
\usepackage{microtype}      % microtypography
\usepackage{graphicx}
\usepackage{subcaption}
\usepackage{natbib}
\usepackage{doi}
\usepackage{color}
\usepackage{amsmath}
\usepackage{mathtools, cuted}
\usepackage{etoolbox}
\usepackage{changepage}
\usepackage{pdflscape}
\usepackage{multicol}
\usepackage{balance}
\usepackage{wrapfig}
\usepackage{array}
\usepackage{soul}
\usepackage{bm}
\usepackage{enumitem}
\usepackage[symbol]{footmisc}
\usepackage[subtle]{savetrees}
\usepackage{longtable}

\usepackage{tikz}

\newcolumntype{L}[1]{>{\raggedright\arraybackslash}p{#1}} 

\AfterEndEnvironment{strip}{\leavevmode}

\newcommand{\JEmbeddingVFiveText}{\href{https://huggingface.co/collections/jinaai/jina-embeddings-v5-text}{\texttt{jina-embeddings-v5-text}}}
\newcommand{\JEmbeddingVFiveTextSmall}{\href{https://huggingface.co/jinaai/jina-embeddings-v5-text-small}{\texttt{jina-embeddings-v5-text-small}}}
\newcommand{\JEmbeddingVFiveTextNano}{\href{https://huggingface.co/jinaai/jina-embeddings-v5-text-nano}{\texttt{jina-embeddings-v5-text-nano}}}
\newcommand{\JEmbeddingVFiveTextSmallShort}{\href{https://huggingface.co/jinaai/jina-embeddings-v5-text-small}{\textcolor{qwen-color}{\texttt{j-v5-text-small}}}}
\newcommand{\JEmbeddingVFiveTextNanoShort}{\href{https://huggingface.co/jinaai/jina-embeddings-v5-text-nano}{\textcolor{eurobert-color}{\texttt{j-v5-text-nano}}}}

\newcommand{\rom}[1]{\uppercase\expandafter{\romannumeral #1\relax}}

\def\@fnsymbol#1{\ensuremath{\ifcase#1\or *\or \dagger\or \ddagger\or
   \mathsection\or \mathparagraph\or \|\or **\or \dagger\dagger
   \or \ddagger\ddagger \else\@ctrerr\fi}}
\newcommand{\ssymbol}[1]{^{\@fnsymbol{#1}}}

\definecolor{darkgreen}{rgb}{0.0, 0.5, 0.1}

\definecolor{qwen-color}{RGB}{128,0,128}
\definecolor{eurobert-color}{RGB}{0,51,153}

\title{\JEmbeddingVFiveText: Task-Targeted Embedding Distillation}

\author{xxx$^*$, xxx$^*$, \\
Jina AI GmbH, Prinzessinnenstraße 19--20, 10969 Berlin, Germany \\
\texttt{research@jina.ai}
}

\author{Mohammad Kalim Akram$^*$, Saba Sturua$^*$, Nastia Havriushenko$^*$, \\ \textbf{Quentin Herreros$^*$}, \textbf{Michael G\"unther$^*$}, \textbf{Maximilian Werk$^*$},  \textbf{Han Xiao} \\ Jina by Elastic \\ \texttt{research@jina.ai}}
\date{2024/02/26}

\hypersetup{
pdftitle={Jina Embeddings V4: Universal Embeddings for Multimodal Multilingual Retrieval},
pdfauthor={xxx}
pdfkeywords={Embeddings, Multilingual Models, Token length, Semantic Textual Similarity, Information Retrieval, Text Retrieval},
}

\begin{document}
\maketitle

\def\thefootnote{*}\footnotetext{Equal contribution.}\def\thefootnote{\arabic{footnote}}

\begin{abstract}
Text embedding models are widely used for semantic similarity tasks, including information retrieval, clustering, and classification.
General-purpose models are typically trained with single- or multi-stage processes using contrastive loss functions.
We introduce a novel training regimen that combines model distillation techniques with task-specific contrastive loss to produce compact, high-performance embedding models.
Our findings suggest that this approach is more effective for training small models than purely contrastive or distillation-based training paradigms alone.
Benchmark scores for the resulting models, \texttt{jina-\allowbreak{}embeddings-v5-\allowbreak{}text-small} and \texttt{jina-\allowbreak{}embeddings-v5-\allowbreak{}text-nano}, exceed or match the state-of-the-art for models of similar size.
\texttt{jina-\allowbreak{}embeddings-v5-\allowbreak{}text} models additionally support long texts (up to 32k tokens) in many languages, and generate embeddings that remain robust under truncation and binary quantization.
Model weights are publicly available, hopefully inspiring further advances in embedding model development.
\end{abstract}

\section{Introduction}
\label{sec:introduction}
Information retrieval (IR) systems increasingly rely on text embedding models as first-stage retrievers, replacing or augmenting traditional methods.
These models map queries and documents into a shared dense vector space, enabling efficient retrieval via nearest-neighbor search.
These dense embeddings see use in a wide array of IR applications, including web search, question-answering, and retrieval-augmented generation, as well as other purposes like recommendation systems, clustering, classification and quantification of semantic similarity.

The prevailing architecture for embedding models is a transformer architecture augmented with a pooling layer, first introduced for Sentence-BERT~\cite{reimers2019sentence}.
Recent models, like Qwen3Embeddings~\cite{zhang2025qwen3} and Embedding-Gemma~\cite{vera2025embeddinggemma}, are trained using contrastive learning.
Alternatively, knowledge distillation provides an efficient mechanism for training small models to mimic the behavior of one or more teacher models, as exemplified by the Jasper model~\cite{zhang2024jasper}.

This work combines model distillation with task-specific contrastive loss training, demonstrating that (1) distillation outperforms naive contrastive training, (2) our combined approach leads to further improvements compared to a pure distillation-based approach, and (3) the resulting models perform  on the MTEB benchmarks~\cite{enevoldsenmmteb} on-par with or better than recent models with comparable sizes.

Specifically, this work's contributions are:
\begin{itemize}
    \item \emph{Training Method:} We introduce a new training method that combines distillation with task-specific, specialized training objectives
    \item \emph{Empirical Analysis of Distillation Methods:} We present a comparative analysis of different distillation methods for embedding models. 
    \item \emph{Model Release:} We have released the resulting model weights to the public\footnote{\url{https://huggingface.co/collections/jinaai/jina-embeddings-v5-text}} in order to foster advances in the field.
\end{itemize}

\section{Related Work}
\label{sec:related_work}

Related work spans work about distilling language models in general, research into distillation specifically for embedding models, and contrastive multi-task learning.

\subsection{Language Model Distillation}
Model distillation is an approach to creating compact language models that has been used to create models like DistilBERT~\cite{sanh2019distilbert}. Distillation uses specialized loss functions to align a “student” model with a “teacher.” For DistilBERT, this means one function to align their outputs, and one to align the hidden layers using cosine loss.
Alternatively, MiniLM models~\cite{wang2020minilm} are distilled by mimicking the self-attention behavior of the parent model.
TinyBERT~\cite{jiao2020tinybert} uses a pre-trained version of BERT during pre-training and a fine-tuned version for fine-tuning.
\citet{chen2021simplified} follow up on this work by developing a reranker model using the same technique with additional labeled data.

\subsection{Embedding Model Distillation}
Early approaches~\cite{hofstatter2020improving,menon2022defense} to the distillation of embedding models focused on aligning new models with the similarity scores of teacher models. \citet{kim2023embeddistill} employ a projection layer to align teacher and student embedding spaces and perform distillation on the embeddings directly.
\citet{yang2024translate,musacchio2025xvlm2vec} train cross-lingual dense retrieval models using machine translation.
\citet{zhang2024jasper} introduce techniques for multi-teacher distillation, using both embedding alignment and score-based distillation methods, applied over multiple training stages.
\citet{formontlearning} add a Gaussian kernel-based loss component for multi-teacher distillation. This appears to improve performance for embedding-based distillation with a projection layer setup.
Also, \citet{zhang2025jasper} recently proposed an approach that consists of a distillation and a contrastive training stage.
Unlike our method, it only fine-tunes an existing embedding model and does not address differences in optimization methods for different task types.

% https://ojs.aaai.org/index.php/AAAI/article/view/29947 (maybe include this one)

\subsection{Task-Specific Embedding Training}

Researchers have also proposed a variety of techniques to train embedding models to jointly optimize for different tasks and thereby resolve task conflicts.

Joint optimization to support multiple target domains commonly involves combinations of loss functions~\cite{wang2014knowledge, chen-etal-2024-m3} or varying the training objective during training~\cite{mohr2024multi}. 
Additionally, generating multiple models via task-specific fine-tuning and then merging their weights using “model soup” methods has proven productive~\cite{vera2025embeddinggemma}.

Instruction tuning has been proposed to resolve task conflicts in both text~\cite{su2023one} and image~\cite{zhang2024magiclens} retrieval models.
Instructions enable fine-grained manual adjustments to improve embedding performance for specific domains and task types.
However, achieving strong performance with hand-crafted instructions requires additional labeling effort from practitioners.
Alternatively, LoRA adapters allow task-specific adaptations to be trained independently and have also been shown to resolve task conflicts effectively~\cite{sturua2025jina}.

\section{Model Architecture}
\label{sec:model-architecture}
% %

\begin{figure}[t]
\centering
\includegraphics[width=\linewidth]{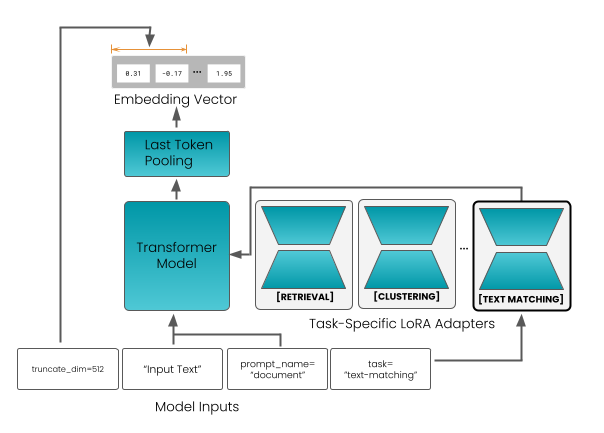}
\caption{Architecture of \JEmbeddingVFiveText{}. 
}
\label{img:model-architecture}
% in https://docs.google.com/presentation/d/1D5yK3Pqv4lGxiDlfOpEvPB6_obTNV4N866YcOT4QhJQ/edit?usp=sharing
\end{figure}
\begin{table}[t]
\small
\setlength{\tabcolsep}{2pt}
\centering
\caption{Attributes of the Base Models and the Resulting Embedding Models}
\label{tab:model_attributes}
\begin{tabular}{lcccccc}
\toprule
\textbf{Model} &
\multicolumn{2}{c}{\textbf{Parameters}} &
\textbf{RoPE} &
\textbf{Max.} &
\textbf{Emb.} \\
\cmidrule(lr){2-3}
\textbf{Name} &
\textbf{Base} &
\textbf{LoRA} &
$\boldsymbol{\theta}$ &
\textbf{Tokens} &
\textbf{Dim.} \\
\midrule
\JEmbeddingVFiveTextSmallShort & 596M & 4$\times{}$20.2M  & 3.5M & 32K & 1024 \\
\JEmbeddingVFiveTextNanoShort & 212M & 4$\times{}$6.7M & 1M & 8K & 768 \\
\midrule
\multicolumn{6}{c}{\textbf{Base Models}} \\
\midrule
\textcolor{qwen-color}{\texttt{Qwen3-0.6B}} & 600M & --  & 1M & 32K & 1024 \\
\textcolor{eurobert-color}{\texttt{EuroBERT-210M}} & 210M & -- & 250K & 8K & 768 \\
\bottomrule
\end{tabular}
\end{table}
Figure~\ref{img:model-architecture} displays the model architecture. It is a transformer model that closely follows the schema of other pre-trained language models~\cite{boizard2025eurobert,yang2025qwen3}. The model translates a text input into a single embedding via last-token pooling, i.e., it uses the embedding of the end-of-sequence token produced by the transformer layers.

Following the approach of \citet{sturua2025jina}, the model includes LoRA adapters to support multiple tasks that are difficult to optimize for jointly. These tasks are: retrieval, semantic similarity, clustering, and classification. Adapters are loaded together with the model weights, and users select the appropriate one at inference time.

To support asymmetric retrieval, \JEmbeddingVFiveText{}
distinguishes between query and document inputs by pre-pending a prefix to the input text -- either "Query:" or "Document:".
Other tasks use a single "Document:" prefix.
Embeddings can also be truncated for downstream efficiency, enabled by using Matryoshka Representation Learning during training~\cite{kusupati2022matryoshka}.

Table~\ref{tab:model_attributes} summarizes the attributes of both embedding models and their underlying backbone models.

\section{Training}
\label{sec:training}
For training our embedding models we use the pre-trained language models \texttt{EuroBERT-210m}~\cite{boizard2025eurobert} for \JEmbeddingVFiveTextNano{} and \texttt{Qwen3-0.6B-Base}~\cite{yang2025qwen3} for \JEmbeddingVFiveTextSmall{} (see Table \ref{tab:model_attributes}).
Both models are multilingual.\footnote{EuroBERT’s training focuses on 15 major European and global languages:  English, French, German, Spanish, Chinese, Italian, Russian, Polish, Portuguese, Japanese, Vietnamese, Dutch, Arabic, Turkish, and Hindi. It also includes some materials in other languages. Qwen3-0.6B-Base lists 119 languages:  \url{https://qwen.ai/blog?id=qwen3} \tiny{(Last Access: 01/27/2026)}.}

Our training method consists of two main stages:

\paragraph{Embedding Distillation:} We use distillation to transfer  knowledge from \texttt{Qwen3-Embedding-4B model}\footnote{\url{https://huggingface.co/Qwen/Qwen3-Embedding-4B} \tiny{(Last Access: 01/27/2026)}}, a much larger, trained embedding model~\cite{zhang2025qwen3}. The goal is to enable a small model to approximate the performance of the larger model without requiring instruction-style prompts or other prompt engineering for embedding generation.
\paragraph{Task-Specific Adapter Training:} 
In this stage, we freeze the model weights and train LoRA adapters for better performance in broad task categories: retrieval, semantic similarity, clustering, and classification.
% Beside prefixes to represent asymmetry in retrieval (see Section~\ref{sec:model-architecture}) no other task-specific instructions are used.

% Before training, model weights are initialized to match \texttt{Qwen/Qwen2.5-VL-3B-Instruct}. The multi-vector projection layer and LoRA adapters are randomly initialized. The weights of the backbone model are not modified during the training process. The LoRA adapters modify the effect of the backbone model layers and the projection layer. Only the adapters are trained.

% Training proceeds in two phases:

% \begin{enumerate}
%     \item A single LoRA adapter is trained using contrasting text pairs and text-image pairs. We use the contrastive InfoNCE~\citep{DBLP:journals/corr/abs-1807-03748} loss function to co-train for both single-vector and multi-vector similarity, as detailed in the section below. No task-specific training is performed at this stage.
%     \item The resulting LoRA adapter is duplicated to create the three task-specific adapters, which are then trained individually with task-specific text triplets and text-image triplets.
% \end{enumerate}

% In both phases of training, we apply {Matryoshka loss}~\cite{kusupati2022matryoshka} to the base loss so that single-vector embeddings from \JEmbeddingVFour{} are truncatable.

\subsection{First-Stage: Embedding Distillation}
\label{sec:embedding-training}

% The objective of the first stage is to distill general knowledge about similarity and relatedness of text values from a larger embedding model.
% We use the \texttt{Qwen3-Embedding-4B model}\footnote{\url{https://huggingface.co/Qwen/Qwen3-Embedding-4B} \tiny{(Last Access: 01/27/2026)}} as teacher model which is open-source and performs well across a wide range of tasks~\cite{zhang2025qwen3}.
% %We divide this stage into two phases: \textit{Base Model Training} and \textit{Long-Context Training}.

% \paragraph{Training Data:}
% Our training data encompass sets of text pairs $(q,d)$ that consist of a left side that resembles queries and a right side that resamples documents e.g., title abstract or question-answer pairs.

Distillation requires a “student” model, a “teacher” model, and training data for both to process.
Our training data consists of text pairs $(q,d)$ that consist of a text that functions as a query $q$ and one that functions as a document to retrieve $d$, e.g., title-abstract and question-answer pairs.

The Qwen3 teacher model has been trained to follow instructions when generating embeddings, enabling users to provide relevant extra-textual information, like whether an embedding is to be used as a query or a document, or domain-relevant information like that a text is a scientific abstract or encyclopedia entry.
This enables the model to position the embedding better in its semantic space and improves task performance. However, it leads to ambiguity when we do not know what instructions are empirically most useful and makes it harder for us to transfer knowledge through distillation.
Therefore, we make only minimal use of instructions during distillation.
For the student, we only provide generic query/document prefixes (described in Section~\ref{sec:retrieval-adapter}), and for the teacher, the general instruction: ``Given a web search query, retrieve relevant passages that answer the query’’, which is provided as a default in its sentence transformer configuration.\footnote{\url{https://huggingface.co/Qwen/Qwen3-Embedding-4B/blob/main/config_sentence_transformers.json} \tiny{(Last Access: 02/13/2026)}}
% Therefore, we do not use comprehensive instructions, but rather only add a prefix for the two prompt types (see Section~\ref{sec:model-architecture}).
% For the teacher model, we use the generic instruction provided in its sentence transformer configuration:  ``Given a web search query, retrieve relevant passages that answer the query''.

\subsubsection{Positional Information}
We use rotary positional embeddings (RoPE)~\cite{su2024roformer} to inject positional information during attention calculation.
This technique uses rotation matrices and a parameter $\theta$, which controls the rotation frequencies.
Using a higher $\theta$ at inference time and a lower one during training has been shown to improve performance on texts that are longer than those seen during training~\cite{zhang2024mgte, liu2024scaling}.
Since our training data consists of relatively short texts, but we want the models to perform well on long ones, we train with much smaller $\theta$ values, as seen in Table~\ref{tab:training_hparams}, than the ones we use at inference time, as shown in Table~\ref{tab:model_attributes}.

% \paragraph{Parameters of Rotary Positional Embeddings (RoPE)}
% RoPE~\cite{su2024roformer} injects positional information during attention calculation via rotation matrices.
% A key hyperparameter is the base $\theta$, which controls rotation frequencies.
% Increasing $\theta$ at inference reduces these frequencies and improves extrapolation to sequence lengths longer than those seen during training~\cite{zhang2024mgte, liu2024scaling}.
% Since most training data consists of short sequences, we train with a smaller $\theta$ (Table~\ref{tab:training_hparams}) and use a larger $\theta$ at inference (Table~\ref{tab:model_attributes}).

\subsubsection{Loss Function}
\label{sec:loss_func}

At each training step, we apply the student/teacher model to a batch of pairs $(q,d)$, resulting in two batches of embeddings:
\[
\mathcal{B}_S = \{ (\mathbf{x}^S_i, \mathbf{y}^S_i) \}_{i=1}^{B},\;\mathbf{x}^S_i, \mathbf{y}^S_i \in \mathbb{R}^n
\]
and
\[
\mathcal{B}_T = \{ (\mathbf{x}^T_i, \mathbf{y}^T_i) \}_{i=1}^{B},\;\mathbf{x}^T_i, \mathbf{y}^T_i \in \mathbb{R}^m
\]
The dimensionality of the teacher embeddings $m$ is higher than the dimensionality of the student embeddings $n$.
We use a linear projection layer $\psi : \mathbb{R}^n \to \mathbb{R}^m,\; \psi(\mathbf{z}) = W \mathbf{z} + \mathbf{b}$ to project the student embeddings into the teacher’s embedding space, enabling us to use cosine similarity $\phi$ to determine similarity scores.
Our distillation loss $\mathcal{L}_{\mathrm{distill}}$ is a sum of cosine distances between the two sets of embeddings:
\begin{align}
\label{eq:distill-loss}
\mathcal{L}_{\mathrm{distill}}
=
\sum_{i=1}^{B}
\Bigg(
\sum_{\mathbf{z} \in \{\mathbf{x},\mathbf{y}\}}
\bigl[1 - \phi\bigl(\psi(\mathbf{z}^S_i),\; \mathbf{z}^T_i\bigr)\bigr]
\Bigg)
\end{align}
Theoretically, it is possible to project the teacher embeddings to the dimensionality of the student embeddings instead. However, we found that this is less effective, as shown in Section~\ref{sec:ablation:projection}.

\subsubsection{Training Procedure}
Distillation proceeds in two phases:

\paragraph{General-Purpose Training:}
First, we performed training using a large, diverse collection of text pairs, drawn from over 300 datasets in over 30 languages.
Training is conducted for 50,000 steps with the hyperparameters documented in Table~\ref{tab:training_hparams}.

% \paragraph{General Domain Training}
% % \subsubsection{General purpose training}
% % \subsubsection{Pair Training}
% \label{sec:base-train}
% At first, we target the model to perform well on all kinds of text domains.
% Therefore, we use a large diverse set of text pairs, drawn from over 300 distinct sub-datasets in over 30 languages.
% Training is conducted for 50,000 steps with hyperparameters listed in Table~\ref{tab:training_hparams}.

\paragraph{Long Context Training:}
\label{sec:long-context}
General-purpose training for \JEmbeddingVFiveTextSmall{} produced unsatisfactory performance on long documents, as shown in Table~\ref{tab:longembed}, and we undertook further training on that model to improve long sequence embeddings quality.
This training incorporated a curated collection of materials, including synthetic documents designed to retrieve documents based on specific contents embedded in long, high-density, noisy texts.
It also contained natural long texts, such as book chapters and long-form articles, paired with LLM-generated queries.
This dataset includes multilingual document-query pairs with texts of 1,000 to 4096 tokens, ensuring that long document performance is robust across languages.

% While the initial distillation training ensures robust semantic understanding, we observed low performance on retrieval tasks with long documents (see Table~\ref{tab:longembed}).
% We aim to mitigate such effects and extend the capacity of \JEmbeddingVFiveTextSmall{} on long sequences by additional training. 

% The long-context data mixture incorporates a specialized selection of documents. We include synthetic 
% % "needle-in-a-haystack"
% datasets, which require the model to retrieve specific values embedded within high-density noise; a narrative reasoning corpus consisting of book chapters and long-form articles paired with generated queries; multilingual text pairs where document lengths exceed 1,000 tokens, ensuring cross-lingual robustness.

We also lowered the $\theta$ parameter of the positional embeddings and increased the maximum sequence length. That facilitates smoother interpolation of frequencies across the extended context window, leading to better performance on long texts.
Detailed hyperparameter configurations are stated in Table~\ref{tab:training_hparams}.

\subsection{Second-Stage: Task-Specific Adapters}
\label{sec:supervised_fine_tuning}

We froze the weights in the distillation-trained model to train the LoRA adapters for specific tasks.
For each task category, we have a separate adapter.
This avoids problems with conflicting optimization objectives.

In this second stage of training, we used different loss functions and training data for each adapter.
We also re-used the projection layer weights trained in the first stage.

% We instantiate three copies of the pair-trained LoRA adapter and give each specific training for its intended task. Training data and loss functions differ for the three tasks.

% \begin{table}[t!]
% \centering
% \small{
% \begin{tabular}{lp{4cm}}
% \toprule
% Task Name      & Description \\
% \midrule
% \texttt{retrieval}        & Asymmetric embedding of queries and documents for retrieval \\ 
% \texttt{text-matching}            &  Semantic text similarity and symmetric retrieval \\
% \texttt{code}           & Retrieving code snippets \\ 
% \bottomrule
% \end{tabular}
% }
% \caption{Supported tasks of \JEmbeddingVFour{}, each corresponding to a LoRA adapter and trained independently}
% \label{tab:task-value}
% \end{table}

%
\subsubsection{Asymmetric Retrieval Adapter}
\label{sec:retrieval-adapter}

Asymmetric retrieval is based on the insight that queries and retrieval targets are usually very different from each other. Queries are almost always much shorter than the document they’re matched to, and are often worded differently, or use different syntax, like question answering. Consequently, encoding queries and documents differently can yield large improvements in retrieval.

We implement this asymmetry with prefixes, specifically by pre-pending "Query:" to inputs intended to be used as queries and "Document:" to texts intended to be retrieval targets (see Section~\ref{sec:model-architecture}).

Training data for this adapter consists of triplet datasets containing queries, relevant documents, and hard negatives, as well as the long-context datasets described in Section~\ref{sec:long-context}.
For texts in the long-context datasets, the maximum sequence length and batch size were adjusted dynamically in each training step, depending on which dataset was sampled. Detailed hyperparameter values are provided in Appendix~\ref{tab:training_hparams}.

We also use a combination of three loss functions:  

\paragraph{Contrastive Loss:}
We use InfoNCE loss~\cite{oord2018representation} with hard negatives~\cite{karpukhin2020dense}. 
Given a batch of size $B$, let $X = \{\bm{x}_i\}_{i=1}^B$ denote the query embeddings and $Y = \{\bm{y}_i\}_{i=1}^B$ their corresponding relevant document embeddings. 
For each query embedding $\bm{x}_i$, we define a negative set $\mathcal{N}_{x_i}$ consisting of all non-matching in-batch document embeddings and additional mined hard negatives, i.e., semantically related but incorrect documents.
Based on the the temperature-scaled exponential cosine similarity $S(\bm{x}, \bm{y}) = \exp(\phi(\bm{x}, \bm{y}) / \tau)$, the contrastive loss is defined as follows:
\begin{equation}
    \label{eq:infonce_loss}
    \mathcal{L}_{\text{NCE}}^{q \rightarrow d}  = - \frac{1}{B} \sum_{i=1}^{B} \ln \left( \frac{S(\mathbf{x}_i, \mathbf{y}_i)}{S(\mathbf{x}_i, \mathbf{y}_i) + \sum\limits_{\mathbf{n} \in \mathcal{N}_{x_i}} S(\mathbf{x}_i, \mathbf{n}) } \right)
\end{equation}
where $\tau$ is a learnable temperature parameter.

\paragraph{Distillation Loss:}
We retain the same knowledge distillation loss used during the first stage of training (Equation~\eqref{eq:distill-loss}), ensuring that the retrieval adapter preserves the general-purpose embedding quality established by the base model.

\paragraph{Spread-Out Regularizer}
Following \citet{vera2025embeddinggemma}, we apply a global orthogonal regularizer (GOR)~\citep{zhang2017learning} that encourages embeddings to be distributed more uniformly across the embedding space, improving their expressive capacity. This also improves robustness to quantization and enables more efficient retrieval under approximate nearest neighbor (ANN) search. The GOR loss is defined as:

% Split the equation to prevent overflowing
\begin{equation}
\label{eq:spread_out_loss}
\begin{aligned}
\mathcal{L}_\text{GOR} = {} &
\frac{1}{B(B-1)} 
\sum_{\substack{i,j \in \mathcal{B} \\ i \neq j}}
(\mathbf{x}_i^\top \mathbf{x}_j)^2 \\
&+
\frac{1}{B(B-1)} 
\sum_{\substack{i,j \in \mathcal{B} \\ i \neq j}}
(\mathbf{y}_i^{+\top} \mathbf{y}_j^{+})^2
\end{aligned}
\end{equation}
where $\mathbf{x}_i$ and $\mathbf{y}_i^{+}$ denote the query and positive document embeddings, respectively. 

This loss penalizes high pairwise similarity between non-matching embeddings, driving them to behave as if uniformly sampled from the unit sphere.

\paragraph{Combined Objective:}
Similarly to ~\citet{vera2025embeddinggemma}, the final training objective for retrieval is a linear combination of the three loss functions:
\begin{equation}
    \label{eq:retrieval_total_loss}
    \mathcal{L}_{\text{retrieval}} = \lambda_{\text{NCE}} \, \mathcal{L}_{\text{NCE}}^{q \rightarrow d} + \lambda_D \, \mathcal{L}_{\text{distill}} + \lambda_S \, \mathcal{L}_\text{GOR}
\end{equation}
where $\lambda_{\text{NCE}}$, $\lambda_D$, and $\lambda_S$ are scalar weights balancing the three objectives.

The final LoRA adapter averages the weights of the last training checkpoint with an earlier checkpoint, employing model averaging to improve performance and robustness.

% Asymmetric retrieval assigns substantially and qualitatively different embeddings to documents and queries, even if they happen to have the very same text. Having distinct encoding mechanisms for the two often significantly benefits embeddings-based retrieval performance. \citet{sturua2024jina} shows that this can be achieved either by training two separate adapters or by employing two distinct prefixes as proposed in \citet{wang2022text}, so that embedding models can readily distinguish them when they generate embeddings.

% We have used the prefix method for \JEmbeddingVFour{}. Previous work shows little benefit from combining both methods.

\subsubsection{Text Matching (STS) Adapter}
\label{sec:text-matching}

We designed the text-matching adapter for semantic text similarity (STS) tasks, i.e., tasks where both text inputs are treated symmetrically, unlike asymmetric retrieval.
This makes the adapter ideal for use cases like duplicate detection, paraphrase identification, or quantifying the similarity of documents in general.

To achieve better symmetric encoding, this adapter uses only the "Document:" prefix during training and inference.

Accurately capturing semantic similarity requires training data with graded annotations, for which we used STS12~\cite{agirre2012semeval}, SICK~\cite{marelli-etal-2014-sick}, and similar datasets.
Our training data is multilingual, including English, German, Spanish, French, and Japanese, among others.
For less-resourced languages, we have relied on machine-translated versions of existing graded annotated datasets.
High-quality human-annotated STS data is very limited in volume, so we supplemented the training data with text pairs drawn from parallel translations and paired paraphrases of texts.

\paragraph{CoSENT Ranking Loss:}
For a batch $\{(\bm{x_i}, \bm{y_i}, s_i)\}_{i=1}^B$ of $B$ training triplets, where $\bm{x_i}, \bm{y_i} \in \mathbb{R}^d$ are embeddings of two text inputs and $s_i \in \mathbb{R}$ is their ground-truth semantic similarity score. we optimize the following ranking-based objective:
% $$\mathcal{L}_{co}(\mathcal{X}) = \ln \left( 1 + \sum_{\substack{i,j \in \{1,\dots,B\} \\ s_i > s_j}} \exp\left( \frac{\phi(x_j, y_j) - \phi(x_i, y_i)}{\tau} \right) \right)$$
\begin{equation}
\label{eq:cosentloss}
    \mathcal{L}_{\mathrm{co}} =
    \ln\!\Bigg[
        1 +
        \sum_{\substack{i,j \in \{1,\dots,B\} \\ s_i > s_j}}
        \frac{
        e^{\phi(\bm{x_j}, \bm{y_j})} -
        e^{\phi(\bm{x_i}, \bm{y_i})}
        }{\tau'}
    \Bigg]
\end{equation}
This loss function ensures that embedding pairs with higher ground-truth similarity tend to receive higher similarity scores than less ground-truth similarity. By aggregating ranking constraints across the batch, it performs a listwise optimization that aligns model-predicted similarities with the ground-truth ordering indicated by human-provided scores. The temperature parameter $\tau' > 0$ controls the smoothness of the objective.

\paragraph{Combined Objective and Distillation:}
To optimize the adapter, we employ a hybrid strategy. 
During each training step, a batch is sampled from a dataset that either contains annotated similarity scores or pairs or triplets without scores. If scores are available, we use the CoSENT loss $\mathcal{L}_{co}$ described above. If the dataset contains unscored pairs and triplets, we use a combination of InfoNCE loss $\mathcal{L}_{\text{NCE}}^{q \rightarrow d}$ and the knowledge distillation loss $\mathcal{L}_{distill}$ as described in Section~\ref{sec:loss_func}:

\begin{equation}
\label{eq:sts_total_loss}
\mathcal{L}_{\text{sts}} =
\begin{cases}
\mathcal{L}_{\text{co}}, 
& \text{if has scores} \\[6pt]
\lambda_{\text{NCE}}\,\mathcal{L}_{\text{NCE}}^{q \rightarrow d}
+ \lambda_{\text{D}}\,\mathcal{L}_{\text{distill}},
& \text{otherwise}
\end{cases}
\end{equation}
For unranked pairs or triplets, we set the weight ratio $\lambda_{nce}:\lambda_{d}$ to $1:2$. This makes sure that the adapter preserves the high-quality semantic features of the teacher model while learning to do symmetric matching. 
For parallel datasets lacking explicit negatives, we use in-batch negatives. 

This switching logic allows the model to benefit from the precision of human-annotated scores while remaining robust through large-scale distillation and contrastive learning.

\subsubsection{Clustering Adapter}
\label{sec:cluster-adapter}

While retrieval tasks require distinguishing documents that are relevant from documents that are only related to a query, clustering tasks require an embedding model to group related documents near each other.
This use is different enough to merit a separate adapter for this task.

As documented in Section~\ref{sec:embedding-training}, the initial distillation training stage uses a generic instruction for the teacher model. We found this to be distinctly suboptimal for clustering tasks. (See Table~\ref{tab:mteb_multilingual_v2_clustering}). 
To solve this problem, we did new distillation training, following the approach in Section~\ref{sec:embedding-training} and using the distillation loss in Equation~\eqref{eq:distill-loss}, but with a clustering-specific instruction for the teacher model: ``Identify the topic or theme of the given document:''.

We trained on pairs of texts derived from sources that are typically used for clustering tasks, e.g., titles and descriptions of news articles.
All texts receive the prefix ``Document:'' when presented to the the student model.
We detail the hyperparameters in Table~\ref{tab:training_hparams}.

\subsubsection{Classification Adapter}
\label{sec:class-adapter}

Classification is a common use case for embeddings, encompassing document categorization, sentiment analysis, intent recognition, and recommendation systems. This can involve embeddings that encode fine-grained semantic information.

Our training data comprises standard classification datasets, including multilabel data, which we converted to single-label format. All datasets consist of text-label pairs, which we transformed into a triplet format: each sample includes one "anchor", one "positive" item that shares the same label as the anchor, and seven "negative" items with different labels. Random selection determined which items from the labeled dataset were deemed anchors, positives, and negatives.
% Preliminary tests for semantic mining were conducted but yielded inconclusive results.

% For both the 0.2B and 0.6B models, we optimized the contrastive loss \eqref{eq:infonce_loss}, adapted for supervised learning. For each anchor text, the target is a randomly selected positive text with the same label. To align the representations better, we calculated the loss in both directions—between the anchor and positive, and vice versa—resulting in the final loss:
For both \JEmbeddingVFiveTextSmall{} and \JEmbeddingVFiveTextNano{}, we used the contrastive loss from Equation \eqref{eq:infonce_loss}.
To adapt it for supervised learning, we use pairs $(q,p)$ of an anchor text and a randomly selected target with the same label.
We optimize with a bi-directional loss function that aligns the representations:

\begin{equation}
\mathcal{L} = \mathcal{L}_{\text{NCE}}^{q \rightarrow d} + \mathcal{L}_{\text{NCE}}^{d \rightarrow q}
\end{equation}

% \begin{equation}
% \mathcal{L}_{\text{NCE}}^{q \rightarrow d}s =\mathcal{L}(anc,pos)+\mathcal{L}(pos,anc)
% \end{equation}

% For $\mathcal{L}_{\text{NCE}}^{q \rightarrow d}"$, the set $\mathcal{N}_{x_i}$ includes all other positives and negatives in the batch, or all other anchors and negatives in the batch. Note that negatives having the same label as the anchor/positive are masked.
For $\mathcal{L}_{\text{NCE}}^{q \rightarrow d}$, the set $\mathcal{N}_{x_i}$ includes all other positives and negatives in the batch.
In contrast, $\mathcal{L}_{\text{NCE}}^{d \rightarrow q}$ uses only in-batch negatives.

We also added a relational knowledge distillation regularizer~\citep{park2019relational} $\mathcal{L}_{\text{r}}$ to prevent feature collapse and enhance the classifier adapter’s zero-shot abilities. The teacher model for this regularization is the base model without the adapter.
%
% \begin{equation}
%     \mathcal{L}_{\text{RKD}} = \frac{1}{M^2} \sum_{i=1}^{M} \sum_{j=1}^{M} \left( \frac{1 - \phi(s_i, s_j)}{\mu_S} - \frac{1 - \phi(t_i, t_j)}{\mu_T} \right)^2
% \end{equation}
\begin{equation}
\mathcal{L}_{\text{r}} =
\sum_{\substack{i,j=1}}^{M}
\frac{1}{M^2}
\left(
\frac{1 - \phi(\bm{s}_i, \bm{s}_j)}{\mu_S}
-
\frac{1 - \phi(\bm{t}_i, \bm{t}_j)}{\mu_T}
\right)^2
\end{equation}

\noindent where $\bm{s}, \bm{t}$ are embeddings from the set of all anchors, positives, and negatives; $M$ is the total number of embeddings (batch size $\times$ 9); and $\mu$ is the scalar mean values of the student and teacher distance matrices. The loss and the regularizer were respectively scaled by weights $\lambda_{\text{NCE}}$ and $\lambda_R$. Hyperparameters are described in Table~\ref{tab:training_hparams}.

\section{Evaluation}
\label{sec:evaluation}

To evaluate our two new models, we apply a variety of embedding evaluation benchmarks to our models, as well as to a selection of comparable models, in order to provide a baseline for comparison. Where evaluation results for those models are reported elsewhere, we took those values instead of redoing all benchmarks.

For general embedding evaluation, we relied on the English MTEB benchmark~\cite{muennighoff2023mteb} and its multilingual version~\cite{enevoldsenmmteb}, with results summarized in Section~\ref{eval:mteb}.
We also conducted a more extensive evaluation of retrieval performance with additional benchmarks outlined in Section~\ref{eval:retrieval}.
To investigate the effects of our novel design choices during the training, we performed ablation studies described in Section~\ref{sec:ablations}, and we tested the robustness of embeddings under truncation in Section~\ref{sec:eval:mrl}.

For comparison, we primarily focus on state-of-the-art multilingual models with similar parameter counts to our models, specifically:
\begin{itemize}
    \item \texttt{jina-embeddings-v3} (\texttt{jina-v3})~\cite{sturua2025jina}
    \item \texttt{snowflake-arctic-embed-l-v2} (\texttt{snowflake-l-v2})~\cite{yu2024arctic}
    \item \texttt{multilingual-e5-large-instruct} (\texttt{mult.-e5-l-instr.})~\cite{wang2024multilingual}
    \item \texttt{KaLM-embedding-multilingual-\allowbreak{}mini-\allowbreak{}instruct-v2.5} (\texttt{KaLM-mini-v2.5})~\cite{zhao2025kalm}
    \item \texttt{voyage4-nano}~\cite{voyage4nano2026}
    \item \texttt{embeddinggemma-300m} (\texttt{Gemma-300M})~\cite{vera2025embeddinggemma}
    \item \texttt{Qwen3-Embedding-0.6B} (\texttt{Qwen3-0.6B}) \cite{zhang2025qwen3}
\end{itemize}

Note that \texttt{Qwen3-Embedding-0.6B} has been trained on the same backbone model as \JEmbeddingVFiveTextSmall{}. 

To determine the influence of instruction-tuning on the performance, we distinguish between \texttt{Qwen3-0.6B (instr.)} and \texttt{Qwen3-0.6B (generic)}. The generic version of the model uses one prefix for each category only, i.e. retrieval, clustering, etc., while the instruction version has an individualized instruction for each dataset. 
% This makes it difficult for us to reproduce the test conditions for \texttt{Qwen3-0.6B (instr.)}, so we can only use the published scores.
%, whereby the instruction version reports scores mentioned on the MTEB leaderboards and the generic version uses a common instruction for each task category (retrieval, clustering, etc.).

We also provide reference comparisons to two much larger models: our teacher model \texttt{Qwen3-Embedding-4B} (\texttt{Qwen3-4B}) \cite{zhang2025qwen3}, and our previous model \texttt{jina-embeddings-v4} (\texttt{jina-v4})~\cite{gunther2025jina}.
Scores published here come from the relevant MTEB learderboards\footnote{\url{https://huggingface.co/spaces/mteb/leaderboard} \tiny{(Last Access: 02/09/2026)}} or our own evaluation if not published elsewhere.

All retrieval tasks were evaluated using nDCG@10, except for Passkey and Needle, which used nDCG@1.
For semantic textual similarity (STS) and summarization tasks, we calculated the Spearman correlation coefficient. 
For clustering tasks, we used the V-measure\footnote{Specifically, the \texttt{scikit-learn} implementation~\cite{pedregosa2011scikit}:  the harmonic mean of homogeneity and completeness, $V = \frac{2hc}{h+c}$. Homogeneity measures cluster purity (each cluster contains mostly one true class), while completeness measures class concentration (each true class is mostly assigned to a single cluster).
} to evaluate the quality of the embeddings.
Classification and reranking tasks were evaluated using accuracy and precision metrics.

\subsection{Performance on MTEB Benchmarks}
\label{eval:mteb}
\begin{table*}[t]
\caption{MTEB (Multilingual, v2) Evaluation Results}
\label{tab:mteb_multi_v2}
\centering
\small
\setlength{\tabcolsep}{2pt}
\begin{tabular}{@{}lccccccccccccc@{}}
\toprule
\textbf{Model} & \textbf{Params} & \textbf{Dim} & \textbf{Avg Tasks} & \textbf{Avg Type} & \textbf{BM} & \textbf{Cls} & \textbf{Clu} & \textbf{IR} & \textbf{MLC} & \textbf{Pair} & \textbf{RR} & \textbf{Ret} & \textbf{STS} \\
\midrule
Qwen3-4B                & 4B    & 2560 & \textbf{69.5} & \textbf{60.9} & \textbf{79.4} & \textbf{72.3} & \textbf{57.1} & \textbf{11.6} & \textbf{26.8} & \textbf{85.1} & \textbf{65.1} & \textbf{69.6} & \textbf{80.9} \\
jina-v4                          & 3.8B  & 2048 & 58.17            & 51.55            & 62.4\textsuperscript{†}          & 55.2\textsuperscript{†}          & 44.0\textsuperscript{†}          & 0.7\textsuperscript{†}            & 19.3\textsuperscript{†}          & 79.3\textsuperscript{†}          & 62.20\textsuperscript{†}            & 66.4          & 74.4          \\
\midrule
Qwen3-0.6B (instr.)              & 596M  & 1024 & 64.3 & 56.0 & 72.2 & 66.8 & 52.3 & \textbf{5.1} & 24.6 & 80.8 & 61.4 & 64.7 & 76.2 \\
Qwen3-0.6B (generic)             & 596M  & 1024 & 61.1 & 54.3 & 72.2 & 58.4\textsuperscript{†} & 49.8\textsuperscript{†} & 3.8\textsuperscript{†}          & 21.1\textsuperscript{†} & 80.8 & 62.2\textsuperscript{†} & 64.2\textsuperscript{†} & 76.2 \\
jina-v3                          & 572M  & 1024 & 58.4 & 50.7 & 65.3 & 58.8 & 45.6 & -1.3         & 18.4 & 79.3 & 57.1 & 55.8 & 77.1 \\
snowflake-l-v2                   & 568M  & 1024 & 57.0 & 50.0 & 64.1 & 57.4 & 42.8 & -2.5         & 18.9 & 76.7 & 63.7 & 58.4 & 70.1 \\
mult.-e5-l-instr.                & 560M  & 1024 & 63.2 & 55.1 & \textbf{80.1} & 64.9 & 50.8 & -0.4 & 22.9 & 80.9 & 62.6 & 57.1 & 76.8 \\
\JEmbeddingVFiveTextSmallShort{} & 677M  & 1024 & \textbf{67.0} & \textbf{58.9} & 69.7 & \textbf{71.3} & \textbf{53.4} & 1.3 & \textbf{42.0} & \textbf{82.9} & \textbf{65.7} & \textbf{64.9} & \textbf{78.9} \\
\midrule
KaLM-mini-v2.5                   & 494M  & 896  & 60.1 & 52.4 & 65.0\textsuperscript{†} & 61.2\textsuperscript{†} & \textbf{53.8}\textsuperscript{†} & -0.6\textsuperscript{†} & 21.0\textsuperscript{†} & 79.1\textsuperscript{†} & 62.4\textsuperscript{†} & 57.9\textsuperscript{†} & 71.9\textsuperscript{†} \\
voyage-4-nano                    & 340M  & 2048 & 58.9 & 52.0 & 64.1\textsuperscript{†} & 58.6\textsuperscript{†} & 45.4\textsuperscript{†} & 3.5\textsuperscript{†}  & 20.1\textsuperscript{†} & 76.3\textsuperscript{†} & 63.1\textsuperscript{†} & \textbf{63.6}\textsuperscript{†} & 73.0\textsuperscript{†} \\
Gemma-300M                       & 308M  & 768  & 61.1 & 54.3 & 64.4 & 60.9 & 51.2 & \textbf{5.6} & 24.8 & 81.4 & 63.2 & 62.5 & 74.7 \\
\JEmbeddingVFiveTextNanoShort{}  & 239M  & 768  & \textbf{65.5} & \textbf{57.7} & \textbf{67.7} & \textbf{69.2} & 52.7 & 0.0 & \textbf{41.3} & \textbf{81.9} & \textbf{64.6} & 63.3 & \textbf{78.2} \\
\bottomrule
\end{tabular}

\medskip
\begin{flushleft}
\textbf{Task Abbreviations:}
Avg Tasks:~Average (Task),
Avg Type:~Average (Task Type),
BM:~Bitext Mining,
Cls:~Classification,
Clu:~Clustering,
IR:~Instruction Reranking,
MLC:~Multilabel Classification,
Pair:~Pair Classification,
RR:~Reranking,
Ret:~Retrieval,
STS:~Semantic Textual Similarity \\
\textsuperscript{†}~(partially) self-evaluated
\end{flushleft}
\end{table*}

Table~\ref{tab:mteb_multi_v2} shows results on the multilingual MTEB (MMTEB) benchmark for \JEmbeddingVFiveTextSmall{} (\JEmbeddingVFiveTextSmallShort{}), \JEmbeddingVFiveTextNano{} (\JEmbeddingVFiveTextNanoShort{}) and other multilingual models.
Scores for individual tasks appear in Appendix~\ref{app:mmteb}.

Compared to other small models, both \JEmbeddingVFiveText{} models achieve the highest average scores in their size category.
The \texttt{Qwen3-4B} model, which we used as the teacher model, still significantly outperforms our models, but it has more than five times as many parameters as \JEmbeddingVFiveTextSmall{} and sixteen times as many as \JEmbeddingVFiveTextNano{}.
\texttt{KaLM-mini-v2.5} achieves slightly better results on clustering tasks than our models, and
\texttt{Voyage-4-nano} has been narrowly trained to focus on retrieval, and has slightly higher benchmark performance than \JEmbeddingVFiveTextNano{}{} in that one category.

\texttt{Qwen3-0.6B} and \texttt{Gemma-300M} also have generally good average MMTEB scores.
Our evaluation of \texttt{Qwen3-0.6B (generic)} with only one instruction defined individually for each task category shows that performance is generally higher when task-level instructions are used, with the exception of reranking tasks.
The differences are most pronounced for classification tasks and less significant for other task categories.
Note that for STS, pair classification, and bitext mining, Qwen does not define task-specific instructions at the individual task level, accordingly, the scores are identical.

\begin{figure}[t]
    \centering
    \includegraphics[width=\linewidth]{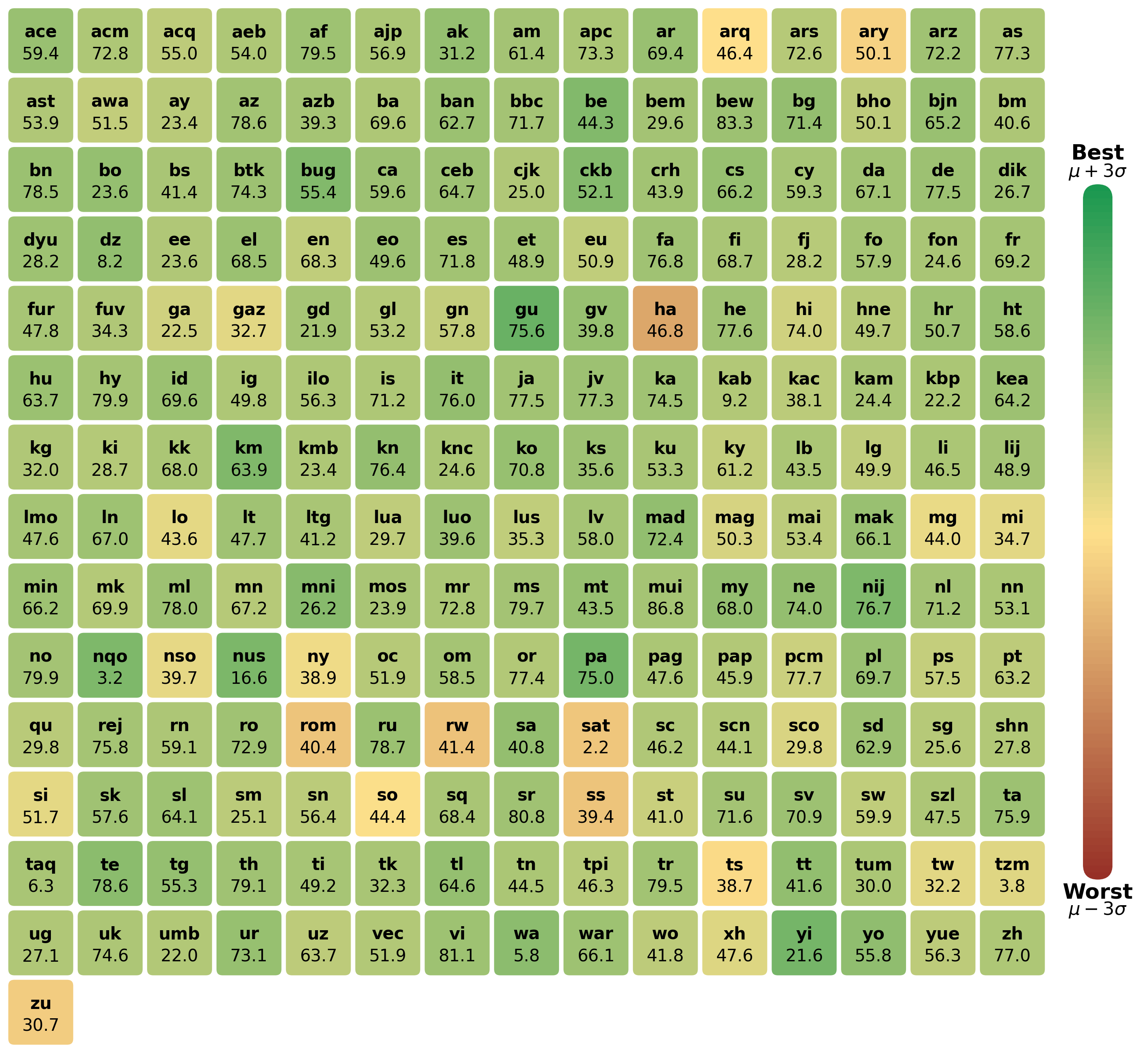}
    \caption{Performance of \JEmbeddingVFiveTextSmallShort{} on different languages on MMTEB compared to other models}
    \label{fig:langs-small}
\end{figure}

Table~\ref{tab:mteb_multi_v2} does not provide insight into language-specific differences in performance, so we conducted separate analyses, calculating average scores for individual languages, for five small multilingual models with published scores for all MMTEB tasks:
\begin{itemize}
    \item \texttt{Gemma-300M}
    \item \texttt{Qwen3-0.6B (instr.)}
    \item \texttt{BGE-M3}~\cite{chen-etal-2024-m3}
    \item \JEmbeddingVFiveTextSmall{}
    \item \JEmbeddingVFiveTextNano{}
\end{itemize}

Figure~\ref{fig:langs-small} presents the average scores per language for \JEmbeddingVFiveTextSmall{} as a heat map, with colors them based on its performance compared to the other four models.\footnote{
Specifically, the color space is mapped to the interval $\mu \pm 3\sigma$ for each individual language.}
Appendix~\ref{app:language-eval} contains heat maps for all five models.

\begin{table*}[t]
\caption{MTEB(eng, v2) Evaluation Results}
\label{tab:mteb_eng_v2}
\centering
\small
\setlength{\tabcolsep}{2pt}
\begin{tabular}{@{}lcccccccccccc@{}}
\toprule
\textbf{Model} & \textbf{Params} & \textbf{Dim} & \textbf{Avg Tasks} & \textbf{Avg Type} & \textbf{Cls} & \textbf{Clu} & \textbf{Pair} & \textbf{RR} & \textbf{Ret} & \textbf{STS} & \textbf{Sum} \\
\midrule
Qwen3-4B                     & 4B    & 2560 & \textbf{74.6} & \textbf{68.1} & \textbf{89.8} & \textbf{57.5} & \textbf{87.0} & \textbf{50.8} & \textbf{68.5} & \textbf{88.7} & \textbf{34.4} \\
jina-v4                    & 3.8B  & 2048 & 65.09   & 60.68   & 74.1\textsuperscript{†} & 45.5\textsuperscript{†} & 83.1\textsuperscript{†} & 48.04\textsuperscript{†}   & 56.2 & 85.9 & 32.0\textsuperscript{†} \\
\midrule
Qwen3-0.6B (instr.)                 & 596M  & 1024 & 70.5 & 64.7 & 84.6 & 54.1 & 84.4 & 48.2 & \textbf{61.8} & 86.6 & \textbf{33.4} \\
Qwen3-0.6B (generic)                  & 596M  & 1024 & 67.0 & 62.0 & 72.0\textsuperscript{†} & 51.8\textsuperscript{†} & 84.4 & 46.2\textsuperscript{†} & 59.8\textsuperscript{†} & 86.6 & \textbf{33.4} \\
jina-v3                    & 572M  & 1024 & 65.7 & 62.6 & 85.8 & 47.4 & 84.0 & 47.9 & 54.3 & 85.8 & 32.9 \\
snowflake-l-v2              & 568M  & 1024 & 63.6 & 59.8 & 73.4 & 44.4 & 83.0 & 47.5 & 58.6 & 78.1 & 33.8 \\
mult.-e5-l-instr.         & 560M  & 1024 & 65.5 & 61.2 & 75.5 & 49.9 & \textbf{86.2} & 48.7 & 53.5 & 84.7 & 29.9 \\
\JEmbeddingVFiveTextSmallShort{}         & 677M  & 1024 & \textbf{71.7} & \textbf{65.6} & \textbf{90.4} & \textbf{54.7} & 85.0 & \textbf{49.4} & 60.1 & \textbf{88.1} & 31.8 \\
\midrule
KaLM-mini-v2.5 & 494M  & 896  & \textbf{71.3} & \textbf{65.3} & \textbf{90.5} & \textbf{58.1} & \textbf{86.6} & 47.4 & 58.5 & 84.8 & 31.2 \\
voyage-4-nano                           & 340M  & 2048 & 63.3 & 58.8 & 73.9\textsuperscript{†} & 46.9\textsuperscript{†} & 83.0\textsuperscript{†} & 47.7\textsuperscript{†} & 52.3\textsuperscript{†} & 81.6\textsuperscript{†} & 26.2\textsuperscript{†} \\
Gemma-300M                   & 308M  & 768  & 69.7 & 65.1 & 87.6 & 56.6 & 87.3 & 47.4 & 55.7 & 83.6 & \textbf{37.6} \\
\JEmbeddingVFiveTextNanoShort{} & 239M  & 768 & 71.0 & 65.2 & 89.7 & 53.5 & 84.7 & \textbf{49.2} & \textbf{58.8} & \textbf{88.3} & 31.9 \\
\bottomrule
\end{tabular}
\medskip
\begin{flushleft}
\textbf{Task Abbreviations:}
Avg Tasks:~Average (Task),
Avg Type:~Average (Task Type),
Cls:~Classification,
Clu:~Clustering,
Pair:~Pair Classification,
RR:~Reranking,
Ret:~Retrieval,
STS:~Semantic Textual Similarity,
Sum:~Summarization \\
\textsuperscript{†}~(partially) self-evaluated
\end{flushleft}
\end{table*}

Table~\ref{tab:mteb_eng_v2} presents the English MTEB benchmark results for all included models.
For results on individual tasks, see Appendix~\ref{app:en-mteb}.

Here, \JEmbeddingVFiveTextSmall{} achieves the highest average score among the small multilingual models, but a lower score than \texttt{Qwen3-4B}.
When examining specific task categories, \texttt{Qwen3-0.6B} achieves slightly better retrieval performance when used with instructions, and \texttt{multilingual-e5-large-instruct} obtains the best results on pair classification tasks.
Using \texttt{Qwen3-0.6B} without individual instructions for each task leads to a similar loss of performance for English benchmarks as was observed for MMTEB.

Among models with fewer than 500M parameters, \texttt{KaLM-mini-v2.5} achieves the highest average scores, only slightly better than \JEmbeddingVFiveTextNano{}, despite having more than twice as many parameters.
\JEmbeddingVFiveTextNano{} achieves higher performance than all other models under 0.5B parameters in retrieval, reranking, and STS tasks. We note that Gemma-300M has the highest overall performance on summarization. %, but only beats \JEmbeddingVFiveTextNano{} by a very small amount.

\subsection{Performance on Various Retrieval Benchmarks}
\label{eval:retrieval}

\begin{table*}[t]
\caption{Retrieval Benchmark Results}
\label{tab:retrieval_benchmarks}
\centering
\small
\setlength{\tabcolsep}{3pt}
\begin{tabular}{@{}lcccccccc@{}}
\toprule
\textbf{Model} & \textbf{Params} & \textbf{Dim} & \textbf{Avg Tasks} & \textbf{MTEB-M} & \textbf{MTEB-E} & \textbf{RTEB} & \textbf{BEIR} & \textbf{Long} \\
\midrule
Qwen3-4B                & 4B    & 2560 & \textbf{67.95} & \textbf{69.60} & \textbf{68.46} & \textbf{70.77}\textsuperscript{†} & \textbf{61.58} & \textbf{78.82}\textsuperscript{†} \\
jina-v4                 & 3.8B  & 2048 & 63.62    & 66.43 & 56.15 & 66.52 & 53.97\textsuperscript{†}  & 69.88 \\
\midrule
Qwen3-0.6B             & 596M  & 1024 & 61.87 & 64.65 & \textbf{61.83} & 64.21\textsuperscript{†} & 55.52 & \textbf{72.20}\textsuperscript{†} \\
jina-v3                 & 572M  & 1024 & 56.11 & 55.76 & 54.29 & 54.58\textsuperscript{†} & 53.17 & 55.67 \\
snowflake-l-v2          & 568M  & 1024 & 57.59 & 58.36 & 58.56 & 53.95 & 55.22 & 63.74 \\
mult.-e5-l-instr.       & 560M  & 1024 & 54.22 & 57.12 & 53.47 & 54.78 & 52.74 & 41.76 \\
\JEmbeddingVFiveTextSmallShort{}         & 677M  & 1024 & \textbf{63.28} & \textbf{64.88} & 60.07 & \textbf{66.84} & \textbf{56.67} & 66.39 \\
\midrule
KaLM-mini-v2.5          & 494M  & 896  & 56.58 & 57.90 & 58.45 & 56.51\textsuperscript{†} & 55.00\textsuperscript{†} & 43.35\textsuperscript{†} \\
voyage-4-nano           & 340M  & 2048 & \textbf{61.48} & \textbf{63.58}\textsuperscript{†} & 52.30\textsuperscript{†} & \textbf{70.36}\textsuperscript{†} & 49.93\textsuperscript{†} & \textbf{74.93}\textsuperscript{†} \\
Gemma-300M              & 308M  & 768  & 59.66 & 62.49 & 55.69 & 63.75\textsuperscript{†} & 53.69\textsuperscript{†} & 55.29\textsuperscript{†} \\
\JEmbeddingVFiveTextNanoShort{}          & 239M  & 768  & 61.43 & 63.26 & \textbf{58.80} & 64.08 & \textbf{56.06} & 63.65 \\
\bottomrule
\end{tabular}
\medskip
\begin{flushleft}
\textbf{Task Abbreviations:}
Avg Tasks:~Task-level mean across benchmarks,
MTEB-M:~MTEB Multilingual v2,
MTEB-E:~MTEB English v2,
RTEB:~RTEB (Multilingual, Public),
BEIR:~BEIR Retrieval,
Long:~LongEmbed\\
\textsuperscript{†}~(partially) self-evaluated
\end{flushleft}
\end{table*}

To provide a more global view of model performance, we used three additional benchmarks: RTEB (Multilingual)\footnote{This benchmark contains a mixture of publicly-available tasks and additional private tasks. These scores here refer to only the public tasks because we do not have access to the private ones.}~\cite{liu2025rteb}, BeIR~\cite{thakur2beir}, and LongEmbed~\cite{zhu2024longembed}. We summarize the results together with the retrieval scores on the MTEB benchmarks from Section~\ref{eval:mteb} in Table~\ref{tab:retrieval_benchmarks}.
Detailed results for individual datasets are presented in Appendix~\ref{app:retrieval-benchmarks}

In contrast to the MTEB retrieval benchmarks, BeIR contains very large English datasets, demonstrating the models' performance on million document-scale corpora. LongEmbed contains tests on relatively long documents when most benchmarks only contain passages. RTEB's tests emphasize model performance on enterprise use cases.

\JEmbeddingVFiveTextSmall{} achieves the highest task-level average across all retrieval benchmarks among the models tested, outperforming comparably-sized \texttt{Qwen3-0.6B} on three out of five benchmarks. \texttt{Qwen3-0.6B} enjoys stronger scores on MTEB English and LongEmbed, suggesting that it has an advantage on English and long-document retrieval tasks. Both \JEmbeddingVFiveText{} models substantially outperform \texttt{jina-v3}, \texttt{snowflake-L-v2}, and multilingual \texttt{e5-large-instruct}. Among models with under 500M parameters, \JEmbeddingVFiveTextNano{} achieves the best BEIR and MTEB English scores while being the smallest model tested. \texttt{Voyage-4-nano} has a slightly higher task-level average than \JEmbeddingVFiveTextNano{}and significantly higher scores on RTEB and LongEmbed. However, \texttt{voyage-4-nano} is bigger in size and has an embedding dimensionality of 2048 compared to \JEmbeddingVFiveTextNano{}'s 768. \texttt{Gemma-300M} and \texttt{KaLM-mini-v2.5} also achieve competitive results on individual benchmarks but fall behind for the overall average across benchmarks. The \texttt{Qwen3-4B} teacher model unsurprisingly achieves the best results across all benchmarks by a considerable margin.

\subsection{Ablation Studies}
\label{sec:ablations}

We analyzed the effect of key design choices in our training setup through ablation testing.
We focused on several factors that directly influence retrieval performance.
Section~\ref{sec:ablation:training-obj} describes empirical studies on different distillation strategies and Section~\ref{sec:ablation:projection} studies the role of student and teacher projections for aligning embedding spaces during distillation.
Furthermore, in Section~\ref{sec:eval:retrieval-component}, we investigate the influence of the three loss components used to train the retrieval adapter, and in Section~\ref{sec:eval:gor-quantization} how GOR regularization makes the model more robust towards binary quantization.
All ablations are conducted using the \texttt{Qwen3-0.6B-Base} backbone model used for \JEmbeddingVFiveTextSmall{}.

\subsubsection{Comparison of Training Objectives}
\label{sec:ablation:training-obj}

We studied the impact of different training objectives on retrieval performance by comparing three distinct loss functions: InfoNCE $\mathcal{L}_{\text{NCE}}^{q \rightarrow d}$ (see Equation~\eqref{eq:infonce_loss}), embedding-based distillation $\mathcal{L}_{\mathrm{distill}}$ (see Equation~\eqref{eq:distill-loss}), and score-based distillation $\mathcal{L}_{\mathrm{score}}$. All models are evaluated on the MTEB English v2 retrieval benchmark, with nDCG@10 reported across training steps.

\paragraph{Score-based distillation loss:} As an alternative to direct embedding alignment with $\mathcal{L}_{\mathrm{distill}}$, we evaluated a score-based distillation loss that aims to match the distribution of pairwise similarities produced by the teacher and student models. Specifically, we compute the Mean Squared Error (MSE) between the softmax-normalized similarity matrices:
\begin{equation}
\label{eq:pairwise-softmax-mse-mean}
% \mathcal{L}_{\mathrm{score}} = \frac{1}{B} \sum_{i=1}^{B} \sum_{j=1}^{B} \left( p^{S}_{i,j} - p^{T}_{i,j} \right)^2,
\mathcal{L}_{\mathrm{score}} = \sum_{\mathbf{z}\in\{\mathbf{x},\mathbf{y}\}}\frac{1}{B}\sum_{i=1}^{B}\sum_{j=1}^{B}\Bigl(p^{S}_{i,j}(\mathbf{z})-p^{T}_{i,j}(\mathbf{z})\Bigr)^2
\end{equation}
where the probability distributions $p^{\alpha}_{i,j}$ for student model ($S$) and teacher model ($T$) are defined as:
\begin{equation}
% p^{\alpha}_{i,j} = \frac{\exp(\phi(\mathbf{z}^{\alpha}_{i}, \mathbf{z}^{\alpha}_{j}) / \tau)}{\sum_{k=1}^{B} \exp(\phi(\mathbf{z}^{\alpha}_{i}, \mathbf{z}^{\alpha}_{k}) / \tau)}.
p^{\alpha}_{i,j}(\mathbf{z})=\frac{\exp\!\bigl(\phi(\mathbf{z}^{\alpha}_{i},\,\mathbf{z}^{\alpha}_{j})/\tau\bigr)}{\sum_{k=1}^{B}\exp\!\bigl(\phi(\mathbf{z}^{\alpha}_{i},\,\mathbf{z}^{\alpha}_{k})/\tau\bigr)},\;\alpha \in \{S,T\}.
\end{equation}
Here, $\phi$ denotes the cosine similarity and $\tau$ is a temperature hyperparameter. To emphasize the importance of higher similarity scores compared to lower similarity scores, we use temperature-scaled softmax values with $\tau=0.02$.

\begin{figure}[t]
\centering
\includegraphics[
  width=\linewidth,
  trim=0.1cm 0cm 0cm 0cm,
  clip
]{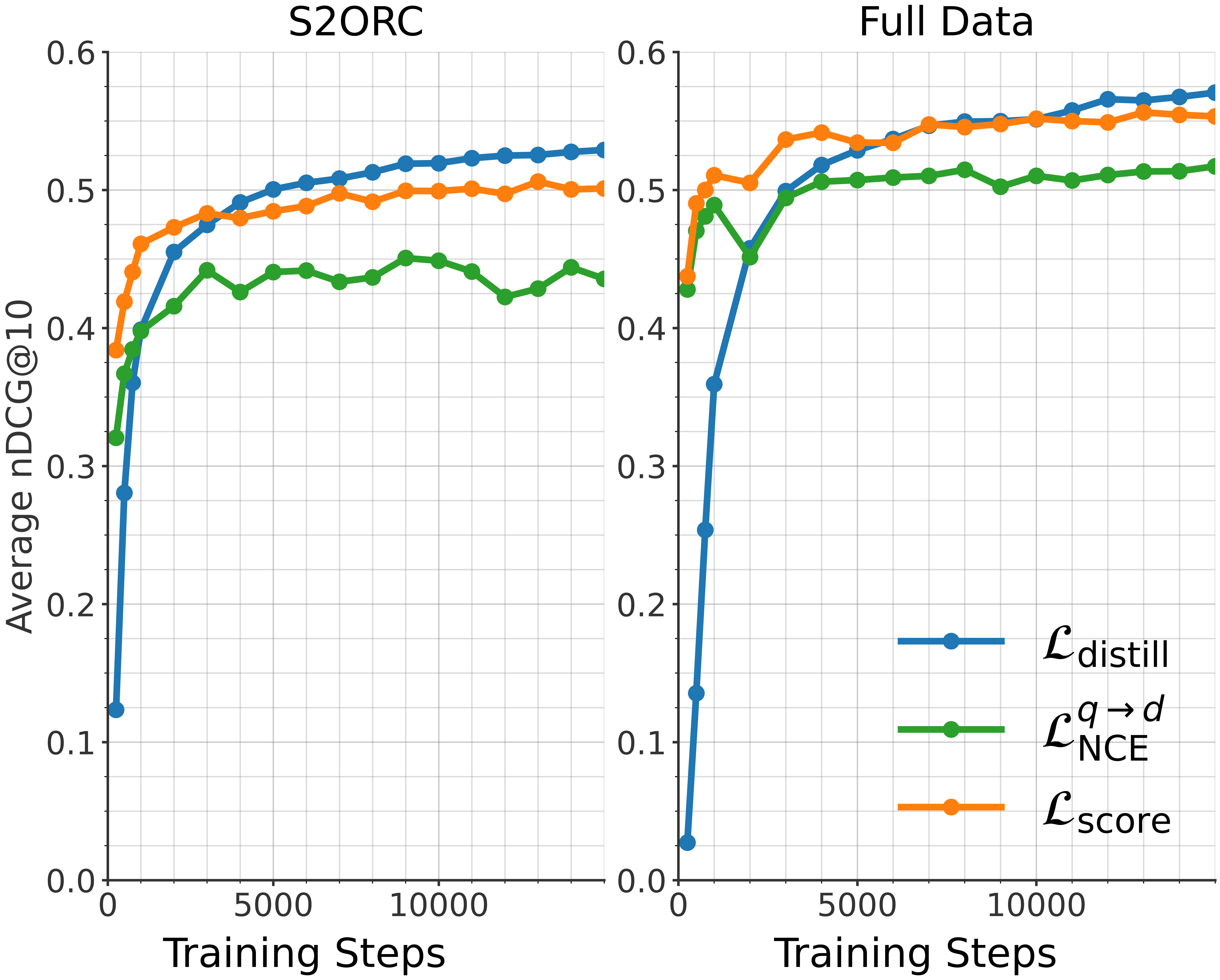}
\caption{Performance comparison of different training objectives. Average nDCG@10 on the MTEB (English, v2) benchmark for \texttt{S2ORC} (left) and the full training data mixture (right).}
\label{fig:distil-obj}
\end{figure}

We conducted these experiments under two data regimes: a filtered version of the \texttt{S2ORC} dataset\footnote{\url{https://huggingface.co/datasets/sentence-transformers/s2orc} \tiny{(Last Access: 02/09/2026)}} and the full data mixture used during the first stage of our training. Detailed hyperparameter configurations for each objective and extended results at different learning rates are provided in Appendix ~\ref{app:learning-rate-ablation}.
% Each run uses two GPUs with a batch size of 512 and a maximum sequence length of 512.
% For embedding-based distillation, use the same learning rate of $1\cdot{}10^{-4}$ as in the first training stage of our models.
% For contrastive training and score-based distillation, we use a lower learning rate of $1\cdot{}10^{-5}$.
% Extended results with different learning rates are provided in Appendix~\todo{add ref}.

% From Figure~\ref{fig:distil-obj}, we observe clear differences in both convergence speed and final performance across training objectives.
% Embedding-based distillation converges more slowly than InfoNCE and score-based distillation, but steadily improves and eventually surpasses both.
% After sufficient training steps, it consistently achieves the highest nDCG@10, indsicating stronger retrieval performance. InfoNCE improves rapidly during the early stages of training but plateaus earlier, while score-based distillation remains competitive initially but shows limited gains in later stages.

% Overall, the results indicate that directly aligning student and teacher embeddings is more effective for training small retrieval models than contrastive supervision or score-level matching alone. Embedding-based distillation not only yields the best final performance, but also exhibits more stable and sustained gains throughout training.

Figure~\ref{fig:distil-obj} illustrates training progress for all three loss functions on the MTEB English v2 retrieval benchmark at nDCG@10. We observe clear differences in both convergence speed and final performance. While $\mathcal{L}_{\text{score}}$ and $\mathcal{L}_{\text{NCE}}$ provide a significantly faster initial increase in scores, they plateau relatively early, with score-based distillation showing very limited progress in later stages. In contrast, embedding-based distillation ($\mathcal{L}_{\text{distill}}$) converges more slowly at the beginning, yet improves steadily and ultimately achieves the highest final retrieval performance in both data regimes. This suggests that while score-level matching is efficient for early alignment, directly aligning student and teacher embeddings provides a stronger and more sustained supervisory signal for long-term refinement.
 
\subsubsection{Projection Layer}
\label{sec:ablation:projection}

\begin{figure}[t]
\centering
\includegraphics[width=\linewidth]{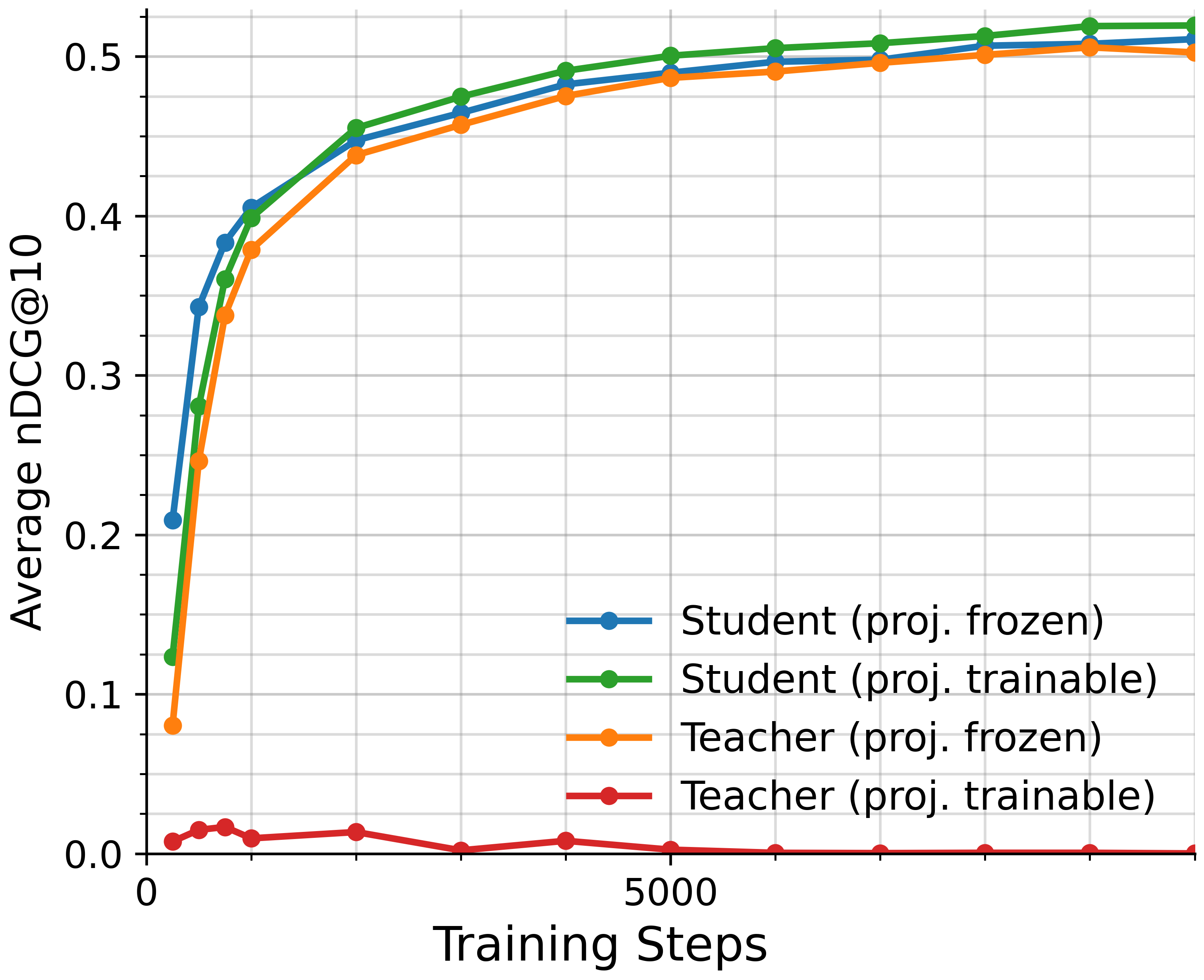}
\caption{Comparison of projection configurations on \texttt{S2ORC}. Performance is measured by average nDCG@10 on MTEB (English, v2).}
\label{fig:eval:projection-ablation}
\end{figure}

We study the effect of projection head placement in embedding-based distillation when aligning models with mismatched embedding dimensions.
All experiments use embedding-based distillation and the \texttt{S2ORC} dataset with the same hyperparameters used for the experiments in Section~\ref{sec:ablation:training-obj}.

We consider two projection strategies to align the embedding spaces: \textit{student projection}, where the student's embeddings are projected into the teacher's embedding space before computing the distillation loss, and \textit{teacher projection}, where the teacher's embeddings are projected into the student's embedding space. In both cases, we evaluated configurations with randomly initialized projections, and with the heads frozen and unfrozen, resulting in four experimental settings.

As shown in Figure~\ref{fig:eval:projection-ablation}, we observe that while teacher projection without freezing simply does not work\footnote{The training probably collapses into a trivial solution.}, all three other configurations perform comparably well. Freezing the student projection leads to faster convergence, and leaving it unfrozen yields the best final results.

\subsubsection{Retrieval Loss Components}
\label{sec:eval:retrieval-component}

% Table 1: Loss Component Ablation
\begin{table}[t]
\centering
\caption{Evaluation of retrieval adapter training losses on MTEB v2 Retrieval subset and public RTEB tasks.}
\label{tab:loss_ablation}
\small
\setlength{\tabcolsep}{3pt}
\begin{tabular}{@{}lcc@{}}
\toprule
\textbf{Loss Configuration} & \textbf{MTEB} & \textbf{RTEB} \\
\midrule
$\mathcal{L}_{\text{NCE}} + \mathcal{L}_{\text{distill}} + \mathcal{L}_\text{GOR}$ & \textbf{64.50} & \textbf{66.45} \\
$\mathcal{L}_{\text{NCE}} + \mathcal{L}_{\text{distill}}$ & 64.21 & 66.16 \\
$\mathcal{L}_{\text{NCE}} + \mathcal{L}_\text{GOR}$ & 64.11 & 66.11 \\
$\mathcal{L}_{\text{distill}} + \mathcal{L}_\text{GOR}$ & 63.49 & 65.05 \\
$\mathcal{L}_{\text{NCE}}$ & 63.38 & 65.14 \\
$\mathcal{L}_{\text{distill}}$ & 63.16 & 64.37 \\
\bottomrule
\end{tabular}
\vspace{2pt}
\end{table}

% Table 2: GOR and Quantization Robustness
\begin{table}[t]
\centering
\caption{Impact of GOR loss on quantization robustness, evaluated on MTEB v2 Retrieval subset and public RTEB tasks.}
\label{tab:gor_quantization}
\small
\setlength{\tabcolsep}{3pt}
\begin{tabular}{@{}lcccc@{}}
\toprule
 & \multicolumn{2}{c}{\textbf{MTEB}} & \multicolumn{2}{c}{\textbf{RTEB}} \\
\cmidrule(lr){2-3} \cmidrule(lr){4-5}
\textbf{Configuration} & \textbf{BF16} & \textbf{Binary} & \textbf{BF16} & \textbf{Binary} \\
\midrule
Full (w/ GOR) & 64.50 & 62.60 (-1.90) & 66.45 & 63.94 (-2.51) \\
w/o GOR & 64.21 & 61.13 (-3.08) & 66.16 & 62.24 (-3.92) \\
\bottomrule
\end{tabular}
\vspace{2pt}
\end{table}

Table~\ref{tab:loss_ablation} presents an ablation study on the components of our retrieval adapter training loss. We systematically remove individual losses from the full combination (Equation~\ref{eq:retrieval_total_loss}) to assess their individual contributions. The results show that combining all three losses yields the best performance across both benchmarks. 

Notably, we show that relying solely on embedding distillation is insufficient, as  $\mathcal{L}_{\text{distill}}$ alone has the lowest scores (63.16 on MTEB, 64.37 on RTEB) of our tested combinations. This validates our two-stage training approach. While $\mathcal{L}_{\text{distill}}$ distillation provides strong initialization in stage 1 training, the addition of task-specific losses ($\mathcal{L}_{\text{NCE}}$ and $\mathcal{L}_\text{GOR}$) in stage 2 is critical for maximizing retrieval performance.

\subsubsection{GOR Loss and Quantization Robustness}
\label{sec:eval:gor-quantization}

In Table~\ref{tab:gor_quantization}, we present the results of training the model with and without the GOR loss component of Equation~\ref{eq:retrieval_total_loss}, both at full-precision (BF16) and binary quantization. At full precision, $\mathcal{L}_\text{GOR}$ contributes only modestly to performance, improving MTEB scores from 64.21 to 64.50 and RTEB from 66.16 to 66.45. However, its benefit becomes evident under quantization. Without $\mathcal{L}_\text{GOR}$, performance degrades over 50\% more on both MTEB and RTEB benchmarks, from -1.90 to -3.08 on MTEB and from -2.51 to -3.92 on RTEB.

This robustness was the goal of GOR regularization: Ensuring fuller use of the available dimensions in embedding space, making the resulting representations less sensitive to information loss.

\subsection{Truncation Robustness of Embeddings}
\label{sec:eval:mrl}

\begin{figure}[t]
\centering
\includegraphics[
  width=\linewidth,
  trim=0.1cm 0cm 0cm 0cm,
  clip
]{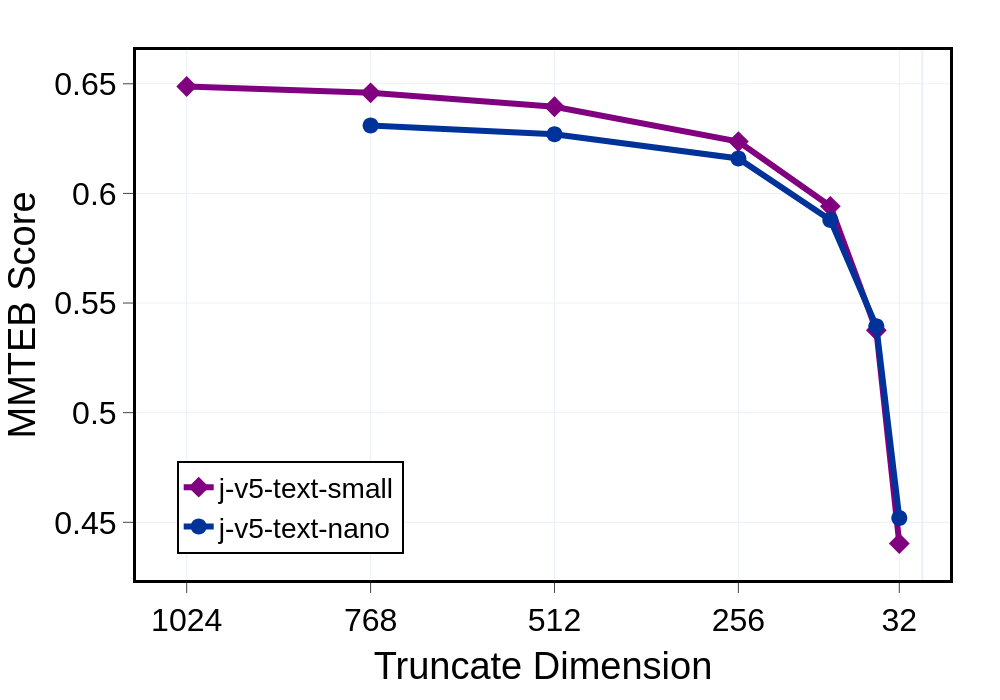}
\caption{Average MMTEB score across reduced embedding dimensions.}
\label{fig:mrl}
\end{figure}

We evaluated the performance of truncated embeddings, the result of using Matryoshka Representation Learning~\cite{kusupati2022matryoshka}.
We progressively reduced to smaller dimensions, in order to assess the efficiency and adaptability of the model’s latent space. Figure~\ref{fig:mrl} shows scores on MMTEB's retrieval benchmarks for embeddings of varying sizes, providing us with a systematic and quantitative analysis of the trade-off between retrieval accuracy and computational efficiency.

Our results show a sizable decline in retrieval performance when the embedding dimensions fall below 256. This aligns with expectations from the Johnson-Lindenstrauss Lemma~\citep{johnson1984lemma}, which establishes theoretical limits on dimensionality reduction while maintaining pairwise distances between data points.

\section{Conclusion}
\label{sec:conclusion}
We have introduced two compact multilingual embedding models \JEmbeddingVFiveTextSmall{} and \JEmbeddingVFiveTextNano{}, and a novel training method for them that combines distillation-based and task-specific training.
We demonstrate through extensive ablation studies that this approach outperforms existing alternatives.
Our models achieve state-of-the-art performance among comparable multilingual embedding models and remain robust under truncation and binary quantization, with only minimal performance degradation in response to large increases in storage and computational efficiency.
To support reproducibility and accelerate future research, we have released the models publicly along with out-of-the-box integration with Sentence Transformers~\citep{reimers2019sentence} and vLLM~\citep{kwon2023efficient}, in addition to multiple quantized variants for llama.cpp~\cite{llama_cpp_ggml_org_2026}.

% We present \JEmbeddingVFour, a state-of-the-art multimodal and multilingual embedding model designed for a wide range of tasks, including semantic text retrieval, text-to-image retrieval, text-to-visually-rich document retrieval, and code search. The model achieves strong performance using single-vector representations and demonstrates even greater effectiveness with multi-vector representations, particularly in visually rich document retrieval.
% \JEmbeddingVFour{} aligns representations across modalities into a single, shared semantic space, sharply reducing structural gaps between modalities compared to CLIP-style dual-tower models, enabling more effective cross-modal retrieval.

% In future work, we plan to further enhance this model’s multilingual capabilities and explore techniques to create smaller, more efficient variants.

\FloatBarrier

\balance

\bibliographystyle{unsrtnat}
\bibliography{references}  %%% Uncomment this line and comment out the ``thebibliography'' section below to use the external .bib file (using bibtex).

\clearpage
\pagenumbering{gobble}
\onecolumn

\appendix
\section{Appendix}
\setcounter{table}{0}
\renewcommand{\thetable}{A\arabic{table}}

\subsection{Hyperparameters}
The following table outlines all hyperparameters used during the various training phases.
For all the LoRA adapters we use a rank of 32 and an alpha value of 32. 

\begin{table}[H]
\centering
\small
\caption{Hyperparameters for the different models and training stages.}
\label{tab:training_hparams}
\setlength{\tabcolsep}{5pt}
\renewcommand{\arraystretch}{1.15}
\begin{tabular}{llcccccl}
\toprule
Stage & Model & Steps & \makecell{Devices \& \\ Batch size} & \makecell{Max. Tokens \\ (Seq. Length)} & LR & $\theta$ & Others \\
\midrule
\multirow{2}{*}{First Stage} 
  & \JEmbeddingVFiveTextSmallShort{} & 50,000 & 8$\times{}$512 & 512 & $1\cdot{}10^{-4}$ & 1M & \\
  & \JEmbeddingVFiveTextNanoShort{} & 50,000 & 8$\times{}$1024  & 512 & $1\cdot{}10^{-4}$ & 250K &  \\
\midrule
% Long-Context Training & \JEmbeddingVFiveTextSmallShort{} & 6,500 & 2$\times{}$64 & 4096 & $1\cdot{}10^{-4}$ & 500K &  \\
Long Context
& \multirow{2}{*}{\centering\JEmbeddingVFiveTextSmallShort{}} 
& \multirow{2}{*}{\centering 6,500} 
& \multirow{2}{*}{\centering 2$\times{}$64} 
& \multirow{2}{*}{\centering 4096} 
& \multirow{2}{*}{\centering $1\cdot{}10^{-4}$} 
& \multirow{2}{*}{\centering 500K} 
& \multirow{2}{*}{\centering} \\
Training & & & & & & & \\
\midrule
\multirow{2}{*}{Asym. Retrieval} 
  & \JEmbeddingVFiveTextSmallShort{} & 8,000  & 2$\times{}$(256 / 64) & 384 / 4096 & $2\cdot{}10^{-5}$ & 1M & $\tau=0.02,
\lambda_D=2,$\\
  & \JEmbeddingVFiveTextNanoShort{} & 8,000  & 2$\times{}$(384 / 96) & 384 / 4096 & $2\cdot{}10^{-5}$ & 250K & $\lambda_{\text{NCE}}=\lambda_S=1$  \\
\midrule
\multirow{2}{*}{Text-Matching} 
  & \JEmbeddingVFiveTextSmallShort{} & 20000 & 1$\times{}$256 & 384 & $5\cdot{}10^{-5}$ & 1M & $\tau=0.02, \tau'=0.05$, \\
  & \JEmbeddingVFiveTextNanoShort{} & 20000 & 1$\times{}$256  & 384 & $5\cdot{}10^{-5}$ & 250K & $\lambda_{\text{NCE}}=1,\lambda_D=2$ \\
\midrule
\multirow{2}{*}{Clustering} 
  & \JEmbeddingVFiveTextSmallShort{} & 20,000 & 1$\times{}$512 & 512 & $1\cdot{}10^{-5}$ & 100K &  \\
  & \JEmbeddingVFiveTextNanoShort{} & 20,000 & 1$\times{}$1024 & 512 & $1\cdot{}10^{-5}$ & 25K &  \\
\midrule
\multirow{2}{*}{Classification} 
  & \JEmbeddingVFiveTextSmallShort{} & 30,000 & 4$\times{}$64 & 512 & $4\cdot{}10^{-4}$ & 3.5M & $\tau=0.02,$ \\
  & \JEmbeddingVFiveTextNanoShort{} & 30,000 & 4$\times{}$128 & 512 & $4\cdot{}10^{-4}$ & 1M &  $\lambda_{\text{NCE}}=1,\lambda_R=20$\\
\bottomrule
\end{tabular}
\end{table}

\subsection{English MTEB Benchmarks}
\label{app:en-mteb}

The following evaluations are computed using the default metrics on the MTEB(eng, v2) benchmark.
We either report results that are stated on the MTEB leaderboard\footnote{\url{https://huggingface.co/spaces/mteb/leaderboard} \tiny{(Last Access: 02/02/2026)}} or self-evaluate them using the mteb package\footnote{\url{https://github.com/embeddings-benchmark/mteb} \tiny{(Last Access: 02/02/2026)}}.

\begin{table}[H]
\centering
\caption{Evaluation Results on MTEB Retrieval Tasks (nDCG@10 [\%])}
\label{tab:mteb_retrieval}
\small{
\setlength{\tabcolsep}{4pt}
\begin{tabular}{lccccccccccc}
\toprule
\textbf{Model} & \textbf{AVG} & \textbf{Arg} & \textbf{CQG} & \textbf{CQU} & \textbf{CFHN} & \textbf{FEV} & \textbf{FiQA} & \textbf{HPQA} & \textbf{SCI} & \textbf{TREC} & \textbf{TOU} \\
\midrule
Qwen3-4B & \textbf{68.46} & \textbf{75.64} & \textbf{71.51} & \textbf{59.60} & \textbf{48.48} & \textbf{92.47} & \textbf{62.65} & \textbf{75.22} & \textbf{31.44} & \textbf{92.92} & \textbf{74.65} \\
jina-v4 & 56.15 & 66.96 & 57.59 & 42.95 & 34.57 & 87.69 & 48.49 & 69.01 & 21.48 & 80.36 & 52.41 \\
\midrule
Qwen3-0.6B (instr.) & \textbf{61.83} & \textbf{70.97} & \textbf{64.14} & \textbf{51.49} & \textbf{43.62} & 88.94 & 46.61 & 67.69 & \textbf{24.41} & \textbf{90.52} & 69.90 \\
Qwen3-0.6B (generic) & 59.77 & 67.48 & 63.83 & 49.80 & 37.61 & 86.66 & 46.83 & 66.90 & 22.77 & 88.80 & 67.01 \\
jina-v3 & 54.29 & 43.29 & 58.02 & 43.52 & 43.14 & 89.90 & 47.35 & 64.70 & 19.92 & 77.74 & 55.28 \\
snowflake-l-v2 & 58.56 & 59.11 & 63.18 & 46.57 & 42.83 & \textbf{92.21} & 45.35 & 68.40 & 20.28 & 83.63 & 64.05 \\
mult.-e5-l-instr. & 53.47 & 58.48 & 63.96 & 44.73 & 23.83 & 75.76 & 48.43 & 64.53 & 19.24 & 82.51 & 53.26 \\
\JEmbeddingVFiveTextSmallShort{} & 60.07 & 65.07 & 62.16 & 49.61 & 41.75 & 90.46 & \textbf{49.63} & \textbf{69.94} & 23.04 & 78.49 & \textbf{70.59} \\
\midrule
KaLM-mini-v2.5 & 58.45 & 60.15 & \textbf{65.52} & \textbf{48.87} & 35.06 & 88.23 & 47.10 & \textbf{71.79} & 21.62 & \textbf{82.98} & 63.23 \\
voyage-4-nano & 52.30 & 58.63 & 60.96 & 46.15 & 22.41 & 68.14 & 50.99 & 63.25 & 21.28 & 77.89 & 53.83 \\
Gemma-300M & 55.69 & \textbf{71.54} & 59.53 & 41.52 & 26.71 & 80.75 & 47.74 & 71.48 & 18.43 & 80.35 & 58.90 \\
\JEmbeddingVFiveTextNanoShort{} & \textbf{58.80} & 65.70 & 61.60 & 47.40 & \textbf{40.03} & \textbf{89.82} & \textbf{47.85} & 69.28 & \textbf{22.60} & 77.60 & \textbf{66.12} \\
\midrule
v5-small stage 1 & 58.52 & 64.05 & 62.00 & 48.42 & 38.19 & 86.46 & 47.53 & 68.45 & 22.28 & 79.48 & 68.36 \\
v5-nano stage 1 & 58.29 & 65.60 & 60.86 & 46.79 & 36.68 & 87.58 & 46.20 & 68.70 & 22.23 & 79.39 & 68.88 \\
\bottomrule
\end{tabular}
}
\begin{flushleft}
\textbf{Tasks}: Avg:~Average over all tasks, Arg:~ArguAna, CQG:~CQADupstackGamingRetrieval, CQU:~CQADupstackUnixRetrieval, CFHN:~ClimateFEVERHardNegatives, FEV:~FEVERHardNegatives, FiQA:~FiQA2018, HPQA:~HotpotQAHardNegatives, SCI:~SCIDOCS, TREC:~TRECCOVID, TOU:~Touche2020Retrieval.v3
\end{flushleft}
\end{table}

\begin{table}[H]
\caption{Evaluation Results on MTEB Reranking Tasks (MAP@1000 [\%]).}
\label{tab:reranking_en_results}
\centering
\small
\setlength{\tabcolsep}{4pt}
\begin{tabular}{l*{3}{c}}
\toprule
\textbf{Model} & \textbf{Avg} & \textbf{MindSmallReranking} & \textbf{AskUbuntuDupQuestions} \\
\midrule
Qwen3-4B & \textbf{50.76} & \textbf{32.71} & \textbf{68.81} \\
jina-v4 & 48.04 & 32.51 & 63.56 \\
\midrule
Qwen3-0.6B (instr.) & 48.18 & 31.23 & 65.13 \\
Qwen3-0.6B (generic) & 46.17 & 31.40 & 60.94 \\
jina-v3 & 47.94 & 30.83 & 65.04 \\
snowflake-l-v2 & 47.47 & 31.59 & 63.35 \\
mult.-e5-l-instr. & 48.74 & 33.07 & 64.41 \\
\JEmbeddingVFiveTextSmallShort{} & \textbf{49.39} & \textbf{32.69} & \textbf{66.08} \\
\midrule
KaLM-mini-v2.5 & 47.42 & 32.45 & 62.39 \\
voyage-4-nano & 47.68 & 32.19 & 63.18 \\
Gemma-300M & 47.43 & 31.90 & 62.96 \\
\JEmbeddingVFiveTextNanoShort{} & \textbf{49.23} & \textbf{32.72} & \textbf{65.73} \\
\midrule
v5-small stage 1 & 47.07 & 32.24 & 61.91 \\
v5-nano stage 1 & 47.64 & 32.43 & 62.85 \\
\bottomrule
\end{tabular}
\end{table}

\begin{table}[H]
\caption{Evaluation Results on MTEB Semantic Textual Similarity Tasks (Spearman correlation [\%]).}
\label{tab:mteb_en_sts}
\setlength{\tabcolsep}{3pt}
\centering
\small
% \begin{tabular}{l*{10}{c}}
% \toprule
% \textbf{Model} & \textbf{Avg} & \textbf{BIO} & \textbf{SICK-R} & \textbf{STS12} & \textbf{STS13} & \textbf{STS14} & \textbf{STS15} & \textbf{STS17} & \textbf{STS22} & \textbf{STSB} \\
% \midrule
% jina-embeddings-v3 & 85.82 & 88.69 & 89.62 & 82.44 & 89.49 & 84.95 & 89.32 & 90.01 & 68.45 & 89.43 \\
% jina-embeddings-v4 & 85.89 & 89.21 & 89.23 & 83.50 & 88.61 & 84.77 & 89.69 & 88.71 & 70.71 & 88.58 \\
% BAAI/bge-m3 & 80.61 & -- & 79.72 & 78.73 & 79.60 & 79.00 & 87.81 & 87.13 & 67.99 & 84.87 \\
% Cohere-embed-English-3 & 82.40 & 83.50 & 81.27 & 74.37 & 85.20 & 80.98 & 89.23 & 90.34 & 68.18 & 88.55 \\
% Cohere-embed-multilingual-v3 & 83.05 & 85.01 & 82.18 & 77.62 & 85.16 & 80.02 & 88.92 & 90.09 & 69.63 & 88.79 \\
% gemini-embedding-001 & 85.29 & 88.97 & 82.75 & 81.55 & 89.89 & 85.41 & 90.44 & 91.61 & 67.97 & 89.08 \\
% multilingual-e5-large & 81.39 & 84.57 & 80.23 & 80.02 & 81.55 & 77.72 & 89.31 & 88.12 & 63.66 & 87.29 \\
% text-embedding-3-large & 81.44 & 84.68 & 79.00 & 72.84 & 86.10 & 81.15 & 88.49 & 90.22 & 66.89 & 83.56 \\
% voyage-3 & 78.59 & 87.92 & 79.63 & 69.52 & 80.56 & 73.33 & 80.39 & 86.81 & 69.60 & 79.53 \\
% voyage-large-2 & 82.63 & 89.13 & 79.78 & 72.94 & 83.11 & 77.21 & 85.30 & 88.77 & -- & 84.78 \\
% voyage-multilingual-v2 & 76.98 & 87.11 & 78.97 & 67.30 & 80.09 & 71.98 & 78.07 & 86.52 & 67.02 & 75.79 \\
% \bottomrule
% \end{tabular}
\begin{tabular}{l*{10}{c}}
\toprule
\textbf{Model} & \textbf{Avg} & \textbf{BIO} & \textbf{SICK-R} & \textbf{STS12} & \textbf{STS13} & \textbf{STS14} & \textbf{STS15} & \textbf{STS17} & \textbf{STS22} & \textbf{STSB} \\
\midrule
Qwen3-4B & \textbf{88.7} & 82.9 & 88.1 & \textbf{86.6} & \textbf{94.4} & \textbf{90.9} & \textbf{93.8} & \textbf{95.1} & \textbf{73.1} & \textbf{93.7} \\
jina-v4 & 85.9 & \textbf{89.2} & \textbf{89.2} & 83.5 & 88.6 & 84.8 & 89.7 & 88.7 & 70.7 & 88.6 \\
\midrule
Qwen3-0.6B (instr.) & 86.6 & 85.5 & 84.8 & 83.0 & \textbf{91.8} & 87.1 & 91.4 & 93.3 & 71.1 & 91.1 \\
Qwen3-0.6B (generic) & 86.6 & 85.5 & 84.8 & 83.0 & \textbf{91.8} & 87.1 & 91.5 & 93.2 & 71.1 & 91.1 \\
jina-v3 & 85.8 & \textbf{88.7} & 89.6 & 82.4 & 89.5 & 85.0 & 89.3 & 90.0 & 68.4 & 89.4 \\
snowflake-l-v2 & 78.0 & 87.2 & 74.0 & 71.2 & 80.4 & 75.1 & 82.9 & 84.5 & 67.9 & 79.2 \\
mult.-e5-l-instr. & 84.7 & 87.5 & 81.7 & 82.5 & 88.1 & 84.8 & 91.0 & 90.3 & 68.2 & 88.4 \\
\JEmbeddingVFiveTextSmallShort{} & \textbf{88.1} & 85.2 & \textbf{91.6} & \textbf{85.1} & 88.7 & \textbf{88.5} & \textbf{92.6} & \textbf{94.9} & \textbf{71.8} & \textbf{94.8} \\
\midrule
KaLM-mini-v2.5 & 84.8 & 84.0 & 83.2 & 81.9 & \textbf{89.5} & 86.0 & 90.3 & 92.3 & 67.2 & 88.9 \\
voyage-nano & 81.6 & 86.6 & 77.9 & 76.0 & 87.4 & 80.2 & 86.2 & 88.9 & 67.0 & 84.5 \\
Gemma-300m & 83.6 & 86.4 & 81.4 & 79.3 & 86.4 & 83.7 & 89.3 & 90.3 & 67.5 & 88.2 \\
\JEmbeddingVFiveTextNanoShort{} & \textbf{88.3} & \textbf{87.5} & \textbf{92.0} & \textbf{85.3} & \textbf{89.5} & \textbf{88.9} & \textbf{92.8} & \textbf{93.7} & \textbf{70.4} & \textbf{94.5} \\
\midrule
v5-small stage 1 & 82.3 & 87.3 & 78.2 & 77.3 & 84.3 & 80.1 & 88.2 & 84.9 & 72.1 & 88.0 \\
v5-nano stage 1 & 83.6 & 87.5 & 81.0 & 80.3 & 86.7 & 81.6 & 88.9 & 90.1 & 68.6 & 87.8 \\
\bottomrule
\end{tabular}
\begin{flushleft}
\textbf{Tasks:} Avg:~Average over all tasks,  BIO:~BIOSSES, STS22:~STS22v2, STSB:~STSBenchmark \\
\end{flushleft}
\end{table}

\begin{table}[H]
\caption{Evaluation Results on MTEB Pair Classification Tasks (Max Average Precision [\%]).}
\label{tab:mteb_pair_classification_en}
\centering
\small
\setlength{\tabcolsep}{4pt}
\begin{tabular}{l*{4}{c}}
\toprule
% \textbf{Model} & \textbf{Avg} & \textbf{SprintDup} & \textbf{TwSem15} & \textbf{TwURL} \\
\textbf{Model} & \textbf{Avg} & \textbf{SprintDuplicateQuestions} & \textbf{TwitterSemEval2015} & \textbf{TwitterURLCorpus} \\
\midrule
Qwen3-4B & \textbf{87.0} & \textbf{96.1} & \textbf{77.8} & \textbf{87.2} \\
jina-v4 & 83.1 & 91.4 & 71.5 & 86.4 \\
\midrule
Qwen3-0.6B (instr.) & 84.4 & 94.1 & 72.3 & \textbf{86.8} \\
Qwen3-0.6B (generic) & 84.4 & 94.1 & 72.3 & \textbf{86.8} \\
jina-v3 & 84.0 & \textbf{97.0} & 70.9 & 84.1 \\
snowflake-l-v2 & 83.0 & 96.5 & 67.0 & 85.4 \\
mult.-e5-l-instr. & \textbf{86.2} & 92.2 & \textbf{79.8} & 86.7 \\
\JEmbeddingVFiveTextSmallShort{} & 85.0 & 96.6 & 72.3 & 86.0 \\
\midrule
KaLM-mini-v2.5 & 86.6 & 96.1 & 77.2 & 86.7 \\
voyage-nano & 83.0 & 93.2 & 70.5 & 86.7 \\
Gemma-300m & \textbf{87.3} & \textbf{97.0} & \textbf{77.9} & \textbf{86.9} \\
\JEmbeddingVFiveTextNanoShort{} & 84.7 & 95.5 & 73.1 & 85.6 \\
\midrule
v5-small stage 1 & 83.4 & 96.4 & 68.4 & 85.5 \\
v5-nano stage 1 & 83.4 & 96.9 & 67.6 & 85.6 \\
\bottomrule
\end{tabular}
\begin{flushleft}
Avg:~Average over  all tasks \\
\end{flushleft}
\end{table}

\begin{table}[H]
\caption{Evaluation Results on MTEB Classification Tasks (Accuracy [\%]).}
\label{tab:mteb_classification_updated}
\centering
\small
\setlength{\tabcolsep}{2pt}
\begin{tabular}{l*{9}{c}}
\toprule
\textbf{Model} & \textbf{Avg} & \textbf{AmzCF} & \textbf{Bnk77} & \textbf{IMDB} & \textbf{MTOP} & \textbf{M-Int} & \textbf{M-Scn} & \textbf{Tox} & \textbf{TwSent} \\
\midrule
Qwen3-4B & \textbf{89.8} & \textbf{93.7} & \textbf{86.3} & \textbf{97.2} & \textbf{97.8} & \textbf{85.0} & \textbf{88.8} & \textbf{91.4} & \textbf{78.4} \\
jina-v4 & 74.1 & 72 & 78.6 & 81.0 & 93.2 & 70.5 & 72.6 & 64.5 & 60.3 \\
\midrule
Qwen3-0.6B (instr.) & 84.6 & 91.5 & 81.0 & 95.4 & 96.0 & 80.4 & 74.1 & 82.1 & \textbf{76.0} \\
Qwen3-0.6B (generic) & 72.0 & 70.1 & 71.6 & 89.6 & 91.0 & 50.0 & 74.4 & 68.0 & 61.5 \\
jina-v3 & 85.8 & 90.9 & 84.1 & 91.9 & 97.5 & 75.2 & 84.1 & 91.3 & 71.4 \\
snowflake-l-v2 & 73.4 & 65.6 & 81.8 & 72.8 & 93.5 & 71.5 & 76.2 & 65.9 & 59.6 \\
mult.-e5-l-instr. & 75.5 & 69.7 & 78.0 & 94.6 & 91.2 & 70.9 & 73.9 & 66.8 & 59.2 \\
\JEmbeddingVFiveTextSmallShort{} & \textbf{90.4} & \textbf{94.3} & \textbf{91.5} & \textbf{95.6} & \textbf{98.9} & \textbf{85.3} & \textbf{92.1} & \textbf{93.7} & 72.1 \\
\midrule
KaLM-mini-v2.5 & \textbf{90.5} & \textbf{94.7} & 90.3 & \textbf{95.9} & 98.7 & 83.2 & 89.3 & 80.1 & \textbf{77.2} \\
voyage-nano & 73.9 & 66.1 & 83.3 & 88.2 & 93.3 & 57.9 & 76.2 & 64.7 & 61.4 \\
Gemma-300M & 87.6 & 90.1 & \textbf{91.5} & 92.9 & \textbf{99.1} & \textbf{85.8} & 91.5 & 82.9 & 66.6 \\
\JEmbeddingVFiveTextNanoShort{} & 89.7 & 93.7 & 90.2 & 94.7 & 98.6 & 84.1 & 91.8 & \textbf{92.8} & 71.5 \\
\midrule
v5-small stage 1 & 75.3 & 75.5 & {84.6} & {82.0} & 94.0 & 56.7 & 77.5 & 69.9 & 62.2 \\
v5-nano stage 1 & 75.0 & 72.7 & 84.3 & 80.7 & {95.0} & {56.8} & 79.1 & 68.1 & 63.1 \\
\bottomrule
\end{tabular}

\vspace{2pt}
\begin{flushleft}
\footnotesize
\textbf{Tasks}: Avg:~Average over all tasks, 
AmzCF:~Amazon Counterfactual Classification, 
Bnk77:~Banking77,
IMDB:~IMDB,
MTOP:~MTOP Domain Classification,
M-Int:~MASSIVE Intent Classification, 
M-Scn:~MASSIVE Scenario Classification, 
Tox:~Toxic Conversations Classification, 
TwSent:~Tweet Sentiment Extraction.
\end{flushleft}
\end{table}

\begin{table}[H]
\caption{Evaluation Results on MTEB Clustering Tasks (V-measure [\%] - see footnote in Section \ref{eval:mteb}).}
\label{tab:mteb_en_clustering}
\setlength{\tabcolsep}{3pt}
\centering
\small
\begin{tabular}{l*{9}{c}}
\toprule
\textbf{Model} & \textbf{Avg} & \textbf{AXP} & \textbf{AXS} & \textbf{BioP} & \textbf{MedP} & \textbf{MedS} & \textbf{SEx} & \textbf{SExP} & \textbf{20News} \\
\midrule
Qwen3-4B & \textbf{57.5} & \textbf{64.8} & \textbf{65.2} & \textbf{50.9} & \textbf{45.3} & \textbf{43.8} & \textbf{77.6} & \textbf{53.5} & \textbf{59.0} \\
jina-v4 & 45.5 & 58.2 & 57.3 & 38.4 & 35.0 & 34.8 & 53.6 & 39.8 & 47.0 \\
\midrule
Qwen3-0.6B (instr.) & 54.1 & 63.7 & \textbf{63.8} & 47.3 & 42.2 & 40.4 & \textbf{71.2} & \textbf{52.1} & 51.7 \\
Qwen3-0.6B (generic) & 51.8 & 63.2 & 61.8 & 45.1 & 40.7 & 39.8 & 66.2 & 45.1 & 52.4 \\
jina-v3 & 47.4 & \textbf{58.9} & 55.9 & 40.0 & 38.2 & 37.2 & 56.7 & 40.9 & 51.5 \\
snowflake-l-v2 & 44.4 & 57.2 & 53.1 & 37.2 & 35.0 & 32.6 & 56.3 & 41.2 & 42.4 \\
mult.-e5-l-instr. & 49.9 & 62.5 & 61.3 & 42.7 & 38.1 & 37.7 & 60.0 & 46.1 & 50.7 \\
\JEmbeddingVFiveTextSmallShort{} & \textbf{54.7} & 65.8 & 62.7 & \textbf{47.7} & \textbf{43.5} & \textbf{41.4} & 70.1 & 52.0 & \textbf{54.3} \\
\midrule
KaLM-mini-v2.5 & \textbf{58.1} & 63.5 & \textbf{61.2} & 50.7 & \textbf{45.6} & \textbf{43.5} & 75.8 & \textbf{51.6} & \textbf{73.1} \\
voyage-nano & 46.9 & 58.2 & 56.4 & 41.2 & 38.3 & 37.1 & 52.3 & 43.1 & 48.5 \\
Gemma-300M & 56.6 & 63.6 & 59.6 & \textbf{52.1} & 44.1 & 41.9 & \textbf{90.9} & 48.9 & 51.3 \\
\JEmbeddingVFiveTextNanoShort{} & 53.5 & \textbf{65.0} & 61.1 & 46.4 & 42.6 & 40.3 & 69.4 & 50.6 & 52.6 \\
\midrule
v5-small stage 1 & 46.5 & 58.2 & 57.0 & 41.8 & 37.8 & 36.7 & 55.4 & 41.5 & 43.8 \\
v5-nano stage 1 & 47.9 & 58.4 & 57.7 & 41.2 & 38.1 & 37.9 & 58.4 & 41.9 & 49.5 \\
\bottomrule
\end{tabular}
\medskip
\begin{flushleft}
\textbf{Tasks:} Avg:~Average over all tasks,
AXP:~ArXivHierarchicalClusteringP2P,
AXS:~ArXivHierarchicalClusteringS2S,
BioP:~BiorxivClusteringP2P.v2,
MedP:~MedrxivClusteringP2P.v2,
MedS:~MedrxivClusteringS2S.v2,
SEx:~StackExchangeClustering.v2,
SExP:~StackExchangeClusteringP2P.v2,
20News:~TwentyNewsgroupsClustering.v2.\\
\end{flushleft}
\end{table}

\subsection{Multilingual MTEB (MMTEB) Benchmarks}
\label{app:mmteb}

The following evaluations are computed using the default metrics on the MTEB(Multilingual, v2) benchmark.
Also here we report results that are stated on the MTEB leaderboard and self-evaluate missing scores.

\begin{table}[H]
\caption{Evaluation Results on MMTEB Retrieval Tasks (nDCG@10\%)}
\label{tab:mmteb_retrieval}
\centering
\small
\setlength{\tabcolsep}{1.6pt}
\begin{tabular}{@{}lccccccccccccccccccc@{}}
\toprule
\textbf{Model} & \textbf{Avg} & \textbf{AI} & \textbf{Arg} & \textbf{Bel} & \textbf{Cov} & \textbf{Hag} & \textbf{PK} & \textbf{LB} & \textbf{MIR} & \textbf{ML} & \textbf{SD} & \textbf{SQA} & \textbf{SO} & \textbf{STC} & \textbf{TC} & \textbf{TR} & \textbf{TW} & \textbf{Wiki} & \textbf{WG} \\
\midrule
Qwen3-4B & \textbf{69.6} & \textbf{81.2} & \textbf{75.6} & \textbf{81.2} & \textbf{87.4} & \textbf{98.8} & \textbf{84.3} & \textbf{95.4} & \textbf{69.5} & \textbf{81.9} & \textbf{31.4} & 20.2 & \textbf{94.3} & 42.3 & \textbf{92.9} & 1.2 & 72.6 & \textbf{91.2} & 51.5 \\
jina-v4 & 66.4 & 50.1 & 67.0 & 74.3 & 80.2 & \textbf{98.8} & 69.8 & 94.8 & 62.9 & 74.9 & 21.5 & \textbf{30.2} & 91.4 & \textbf{58.1} & 80.4 & \textbf{1.3} & \textbf{84.4} & 88.5 & \textbf{67.3} \\
\midrule
Qwen3-0.6B (instr.) & 64.6 & \textbf{79.0} & \textbf{71.0} & 68.7 & \textbf{84.8} & \textbf{98.8} & 84.8 & \textbf{94.5} & 61.2 & 72.8 & \textbf{24.4} & 10.6 & 90.0 & 33.6 & \textbf{90.5} & 1.0 & 60.0 & 87.1 & 50.8 \\
Qwen3-0.6B (generic) & 64.2 & 74.0 & 67.5 & 68.6 & 83.0 & \textbf{98.8} & \textbf{95.3} & 94.0 & 61.4 & 71.2 & 22.8 & 8.3 & 89.3 & 32.2 & 88.8 & 1.1 & 61.1 & 86.8 & 51.2 \\
jina-v3 & 55.8 & 32.8 & 43.3 & 73.4 & 78.5 & 98.7 & 38.0 & 93.4 & 62.6 & 73.4 & 19.9 & 0.7 & 90.8 & 39.2 & 77.7 & 0.6 & \textbf{73.0} & 89.1 & 18.6 \\
snowflake-l-v2 & 58.4 & 22.8 & 59.1 & 74.0 & 78.5 & 98.7 & 77.3 & 93.8 & 66.5 & 73.1 & 20.3 & 5.7 & 86.9 & 19.1 & 83.6 & 1.0 & 44.5 & 90.5 & 55.0 \\
mult.-e5-l-instr. & 57.1 & 29.7 & 58.5 & \textbf{80.9} & 75.8 & 98.7 & 37.8 & 94.3 & 57.7 & 76.2 & 19.2 & \textbf{13.5} & 85.8 & 33.7 & 82.5 & \textbf{1.2} & 36.9 & \textbf{91.6} & 54.3 \\
\JEmbeddingVFiveTextSmallShort{} & \textbf{64.9} & 53.3 & 65.1 & 77.5 & 80.1 & \textbf{98.8} & 80.5 & 94.0 & \textbf{66.6} & \textbf{77.6} & 23.0 & 11.5 & \textbf{93.4} & \textbf{46.3} & 78.5 & 0.9 & 71.6 & 90.6 & \textbf{58.7} \\
\midrule
KaLM-mini-v2.5 & 57.9 & 39.8 & 59.6 & 69.7 & \textbf{83.8} & 98.7 & 38.5 & 94.5 & 63.0 & 72.6 & 17.6 & 5.0 & 91.2 & 37.8 & \textbf{83.2} & \textbf{1.9} & 58.1 & 86.6 & 40.5 \\
voyage-4-nano & \textbf{63.6} & 48.7 & 58.6 & \textbf{80.9} & 79.8 & \textbf{99.0} & \textbf{87.3} & 94.8 & 58.7 & 78.6 & 21.3 & 9.7 & \textbf{94.3} & 40.1 & 77.9 & 1.7 & \textbf{80.5} & \textbf{92.8} & 39.9 \\
Gemma-300M & 62.5 & 37.4 & \textbf{71.5} & 72.4 & 78.9 & 98.9 & 60.8 & \textbf{95.1} & \textbf{66.2} & \textbf{79.0} & 18.4 & \textbf{10.7} & 86.5 & \textbf{46.3} & 80.4 & 1.0 & 72.0 & 90.0 & \textbf{59.4} \\
\JEmbeddingVFiveTextNanoShort{} & 63.3 & \textbf{51.5} & 65.7 & 75.3 & 78.0 & 98.8 & 81.5 & 94.5 & 65.8 & 77.0 & \textbf{22.6} & 6.3 & 92.3 & 33.1 & 77.6 & 1.2 & 74.0 & 90.0 & 53.4 \\
\midrule
v5-small stage 1 & 63.5 & 49.5 & 64.1 & 77.6 & 79.3 & 98.8 & 83.0 & 94.0 & 64.6 & 78.1 & 22.3 & 10.0 & 91.6 & 40.9 & 79.5 & 0.7 & 70.8 & 90.6 & 48.5 \\
v5-nano stage 1 & 62.1 & 47.0 & 65.6 & 75.5 & 77.6 & 98.8 & 83.0 & 94.5 & 63.9 & 77.7 & 22.2 & 6.4 & 89.5 & 31.9 & 79.4 & 1.0 & 71.1 & 89.5 & 44.0 \\
\bottomrule
\end{tabular}
\medskip
\begin{flushleft}
\textbf{Tasks:} Avg:~Average over all tasks, AI:~AILAStatutes, Arg:~ArguAna, Bel:~BelebeleRetrieval, Cov:~CovidRetrieval, Hag:~HagridRetrieval, PK:~LEMBPasskeyRetrieval, LB:~LegalBenchCorporateLobbying, MIR:~MIRACLRetrievalHardNegatives, ML:~MLQARetrieval, SD:~SCIDOCS, SQA:~SpartQA, SO:~StackOverflowQA, TC:~TREC-COVID, STC:~StatcanDialogueDatasetRetrieval, TR:~TempReasonL1, TW:~TwitterHjerneRetrieval, Wiki:~WikipediaRetrievalMultilingual, WG:~WinoGrande
\end{flushleft}
\end{table}

\begin{table}[H]
\caption{Evaluation Results on MMTEB Reranking Tasks (MAP@1000 [\%]).}
\label{tab:reranking_multi_results}
\centering
\small
\setlength{\tabcolsep}{3pt}
\begin{tabular}{l*{7}{c}}
\toprule
\textbf{Model} & \textbf{Avg} & \textbf{Alloprof} & \textbf{RuBQ} & \textbf{T2R} & \textbf{Voyage} & \textbf{WebLINX} & \textbf{Wiki-Multi} \\
\midrule
Qwen3-4B & \textbf{65.08} & \textbf{85.13} & \textbf{72.28} & \textbf{67.27} & \textbf{65.61} & 11.30 & \textbf{88.89} \\
jina-v4 & 62.20 & 78.24 & 70.95 & 64.86 & 58.86 & \textbf{14.00} & 86.28 \\
\midrule
Qwen3-0.6B (instr.) & 61.41 & 80.38 & 65.67 & 67.15 & 57.66 & \textbf{11.60} & 85.99 \\
Qwen3-0.6B (generic) & 62.25 & 79.52 & 69.66 & 66.96 & 61.38 & 10.13 & 85.83 \\
jina-v3 & 57.09 & 72.93 & 65.56 & 65.61 & 50.76 & 9.84 & 77.81 \\
snowflake-l-v2 & 63.67 & 75.84 & 73.75 & 67.57 & 66.63 & 9.05 & 89.18 \\
mult.-e5-l-instr. & 62.61 & 74.68 & 71.66 & 67.12 & 62.48 & 8.71 & \textbf{91.03} \\
\JEmbeddingVFiveTextSmallShort{} & \textbf{65.66} & \textbf{81.39} & \textbf{74.89} & \textbf{68.04} & \textbf{68.84} & 11.33 & 89.46 \\
\midrule
KaLM-mini-v2.5 & 62.36 & 75.99 & 73.59 & 67.45 & 62.10 & 9.71 & 85.31 \\
voyage-4-nano & 63.15 & 78.36 & 73.38 & 65.19 & 59.24 & 10.63 & \textbf{92.07} \\
Gemma-300M & 63.25 & \textbf{79.69} & 71.26 & 67.54 & 61.00 & 10.16 & 89.88 \\
\JEmbeddingVFiveTextNanoShort{} & \textbf{64.63} & 79.67 & \textbf{73.68} & \textbf{67.63} & \textbf{67.02} & \textbf{10.95} & 88.83 \\
\midrule
v5-small stage 1 & 64.72 & 80.60 & 73.19 & 67.69 & 66.36 & 11.01 & 89.45 \\
v5-nano stage 1 & 63.99 & 79.85 & 72.81 & 67.34 & 64.72 & 10.83 & 88.37 \\
\bottomrule
\end{tabular}
\begin{flushleft}
\textbf{Tasks}: Avg:~Average over all tasks, Alloprof:~AlloprofReranking, RuBQ:~RuBQReranking, T2R:~T2Reranking, Voyage:~VoyageMMarcoReranking, WebLINX:~WebLINXCandidatesReranking, Wiki-Multi:~WikipediaRerankingMultilingual.
\end{flushleft}
\end{table}

\begin{table}[H]
\caption{Evaluation Results on MMTEB Semantic Textual Similarity Tasks (Spearman correlation [\%]).}
\label{tab:mmteb_sts}
\setlength{\tabcolsep}{2pt}
\centering
\small
\begin{tabular}{@{}lccccccccccccccccc@{}}
\toprule
\textbf{Model} & \textbf{Avg} & \textbf{Faro} & \textbf{FinP} & \textbf{GSTS} & \textbf{Indic} & \textbf{JSICK} & \textbf{SICK-R} & \textbf{S12} & \textbf{S13} & \textbf{S14} & \textbf{S15} & \textbf{S17} & \textbf{S22} & \textbf{STSB} & \textbf{STSBm} & \textbf{STSES} & \textbf{SRel} \\
\midrule
\textbf{Qwen3-4B} & \textbf{80.9} & \textbf{85.8} & \textbf{34.0} & \textbf{90.0} & \textbf{60.8} & \textbf{88.8} & 88.1 & \textbf{86.6} & \textbf{94.4} & \textbf{90.9} & \textbf{93.8} & \textbf{91.8} & \textbf{73.0} & 86.1 & \textbf{93.7} & 72.8 & \textbf{63.2} \\
jina-v4 & 74.4 & 72.3 & 15.1 & 88.2 & 35.2 & 80.3 & \textbf{89.2} & 83.5 & 88.6 & 84.8 & 89.7 & 85.0 & 71.8 & \textbf{86.6} & 88.6 & \textbf{75.3} & 56.5 \\
\midrule
Qwen3-0.6B (instr.) & 76.2 & 74.3 & 26.3 & 84.9 & 39.0 & \textbf{86.6} & 84.8 & 83.0 & \textbf{91.8} & 87.1 & 91.4 & 85.5 & \textbf{71.8} & 84.6 & 91.1 & 76.9 & 59.4 \\
Qwen3-0.6B (generic) & 76.2 & 74.3 & 26.3 & 84.9 & 39.0 & \textbf{86.6} & 84.8 & 83.0 & \textbf{91.8} & 87.1 & 91.4 & 85.5 & \textbf{71.8} & 84.6 & 91.1 & 76.9 & 59.4 \\
jina-v3 & 77.1 & \textbf{80.8} & 22.4 & 87.9 & \textbf{54.7} & 78.2 & 89.6 & 82.4 & 89.5 & 85.0 & 89.3 & 85.9 & 71.1 & 85.4 & 89.4 & 77.9 & 64.6 \\
snowflake-l-v2 & 70.1 & 70.9 & 22.1 & 77.0 & 47.2 & 81.6 & 74.0 & 71.2 & 80.4 & 75.1 & 82.9 & 74.4 & 68.7 & 72.0 & 79.2 & 78.7 & 66.3 \\
mult.-e5-l-instr. & 76.8 & 80.4 & 25.6 & 85.9 & 53.7 & 82.5 & 81.7 & 82.5 & 88.1 & 84.8 & 91.0 & 86.0 & 69.0 & 83.1 & 88.4 & 77.1 & \textbf{69.2} \\
\textbf{\JEmbeddingVFiveTextSmallShort{}} & \textbf{78.9} & 76.9 & \textbf{33.1} & \textbf{91.0} & 47.8 & 81.3 & \textbf{91.6} & \textbf{85.1} & 88.7 & \textbf{88.5} & \textbf{92.6} & \textbf{87.2} & 71.0 & \textbf{89.9} & \textbf{94.8} & \textbf{81.2} & 60.9 \\
\midrule
KaLM-mini-v2.5 & 71.8 & 64.4 & 22.1 & 83.9 & 15.0 & 79.6 & 83.2 & 81.9 & \textbf{89.5} & 86.0 & 90.3 & 81.3 & \textbf{73.2} & 82.9 & 88.9 & 73.1 & 54.2 \\
voyage-nano & 73.0 & \textbf{73.7} & 20.1 & 81.1 & \textbf{46.3} & 81.6 & 77.9 & 76.0 & 87.4 & 80.2 & 86.2 & 79.1 & 70.8 & 79.2 & 84.5 & 79.3 & \textbf{65.3} \\
Gemma-300m & 74.7 & 65.3 & 25.2 & 84.7 & 43.1 & \textbf{84.4} & 81.4 & 79.3 & 86.4 & 83.7 & 89.3 & 84.4 & 71.2 & 81.6 & 88.2 & \textbf{82.3} & 65.2 \\
\textbf{\JEmbeddingVFiveTextNanoShort{}} & \textbf{78.2} & 71.4 & \textbf{35.0} & \textbf{90.7} & 41.4 & 81.5 & \textbf{92.0} & \textbf{85.3} & \textbf{89.5} & \textbf{88.9} & \textbf{92.8} & \textbf{86.3} & 69.6 & \textbf{89.8} & \textbf{94.5} & 80.1 & 61.9 \\
\midrule
v5-small stage 1 & 74.3 & 75.0 & 18.7 & 83.6 & 44.6 & 86.1 & 78.2 & 77.3 & 84.3 & 80.1 & 88.2 & 84.9 & 72.1 & 82.2 & 88.0 & 80.3 & 64.6 \\
v5-nano stage 1 & 74.5 & 72.0 & 20.7 & 84.2 & 40.3 & 86.0 & 81.0 & 80.3 & 86.7 & 81.6 & 88.9 & 84.3 & 71.8 & 82.5 & 87.8 & 79.8 & 63.7 \\
\bottomrule
\end{tabular}
\begin{flushleft}
\textbf{Tasks:} Avg:~Average over all tasks, Faro:~FaroeseSTS, FinP:~FinParaSTS, GSTS:~ GermanSTSBenchmark, Indic:~IndicCrosslingualSTS, S12:~STS12, S13:~STS13, S14:~STS14, S15:~STS15, S17:~STS17, S22:~STS22.v2, STSB:~STSB, STSBm:~STSBenchmark, SRel:~SemRel24STS
\end{flushleft}
\end{table}

\begin{table}[H]
\caption{Evaluation Results on MMTEB Pair Classification Tasks (Max Average Precision [\%]).}
\label{tab:mmteb_pair_classification}
\centering
\small
\setlength{\tabcolsep}{1.5pt}
\begin{tabular}{l*{12}{c}}
\toprule
\textbf{Model} & \textbf{Avg} & \textbf{ArmPC} & \textbf{CTK} & \textbf{Opus} & \textbf{PawsX} & \textbf{Ppc} & \textbf{RTE3} & \textbf{Sprint} & \textbf{TERRa} & \textbf{TwURL} & \textbf{XNLI} & \textbf{IndoNLI} \\
\midrule
\textbf{Qwen3-4B} & \textbf{85.1} & \textbf{96.3} & \textbf{86.2} & \textbf{95.8} & \textbf{68.9} & \textbf{95.3} & \textbf{90.8} & \textbf{96.1} & \textbf{66.6} & \textbf{87.2} & \textbf{87.2} & \textbf{65.3} \\
jina-v4 & 79.3 & 94.4 & 79.6 & 94.1 & 61.9 & 91.8 & 88.3 & 91.4 & 57.4 & 86.4 & 71.9 & 54.6 \\
\midrule
Qwen3-0.6B (instr.) & 80.8 & 93.1 & 79.2 & 92.9 & 62.2 & 90.5 & 89.0 & 94.1 & 60.7 & \textbf{86.7} & 81.7 & 59.0 \\
Qwen3-0.6B (generic) & 80.8 & 93.1 & 79.2 & 92.9 & 62.2 & 90.5 & 89.0 & 94.1 & 60.7 & \textbf{86.7} & 81.7 & 59.0 \\
jina-v3 & 79.3 & 95.8 & 79.2 & 94.6 & 54.4 & 91.4 & 88.1 & \textbf{97.0} & 59.2 & 84.1 & 73.7 & 54.4 \\
snowflake-l-v2 & 76.7 & 95.9 & 74.5 & 92.7 & 56.6 & 87.1 & 86.2 & 96.5 & 53.8 & 85.4 & 64.3 & 50.7 \\
mult.-e5-l-instr. & 80.9 & \textbf{96.0} & 82.6 & \textbf{95.5} & 55.6 & \textbf{93.5} & 87.9 & 92.2 & 63.9 & \textbf{86.7} & 79.5 & 56.2 \\
\textbf{\JEmbeddingVFiveTextSmallShort{}} & \textbf{82.9} & 94.5 & \textbf{82.7} & 94.1 & \textbf{65.7} & 93.3 & \textbf{89.7} & 96.6 & \textbf{67.3} & 85.9 & \textbf{82.4} & \textbf{60.0} \\
\midrule
KaLM-mini-v2.5 & 79.1 & 92.5 & 74.8 & 92.6 & \textbf{65.0} & 88.0 & 88.2 & 96.1 & 57.5 & 86.7 & 74.3 & 54.8 \\
voyage-nano & 76.3 & \textbf{95.0} & 77.0 & 92.0 & 56.0 & 83.9 & 88.0 & 93.1 & 52.6 & 85.3 & 65.3 & 51.0 \\
Gemma-300m & 81.4 & 92.7 & 79.3 & 93.3 & 57.7 & 90.9 & \textbf{89.7} & \textbf{97.0} & \textbf{65.1} & \textbf{86.9} & \textbf{81.7} & \textbf{61.0} \\
\textbf{\JEmbeddingVFiveTextNanoShort{}} & \textbf{81.9} & 93.5 & \textbf{81.6} & \textbf{94.0} & 62.2 & \textbf{93.9} & 89.4 & 95.5 & 64.6 & 85.6 & 81.4 & 59.6 \\
\midrule
v5-small stage 1 & 77.9 & 93.9 & 75.2 & 93.6 & 59.3 & 87.9 & 86.8 & 96.4 & 57.3 & 85.5 & 69.3 & 51.9 \\
v5-nano stage 1 & 78.0 & 93.8 & 74.6 & 93.7 & 58.7 & 89.1 & 86.6 & 96.9 & 58.1 & 85.6 & 69.0 & 51.9 \\
\bottomrule
\end{tabular}
\begin{flushleft}
\textbf{Tasks}: Avg:~Average over all tasks, ArmPC:~ArmenianParaphrasePC, CTK:~CTKFactsNLI, Opus:~OpusparcusPC, PawsX:~PawsXPairClassification, Ppc:~PpcPC, RTE3:~RTE3, Sprint:~SprintDuplicateQuestions, TERRa:~TERRa, TwURL:~TwitterURLCorpus, XNLI:~XNLI, IndoNLI:~indonli
\end{flushleft}
\end{table}

\begin{table}[H]
\caption{Evaluation Results on MMTEB Bitext Mining Tasks (F1 Score [\%]).}
\label{tab:bitext_mining}
\centering
\small
\setlength{\tabcolsep}{1.5pt}
\begin{tabular}{l*{14}{c}}
\toprule
\textbf{Model} & \textbf{Avg} & \textbf{BUCC} & \textbf{Bible} & \textbf{Bornh} & \textbf{DiaBl} & \textbf{Flores} & \textbf{IN22} & \textbf{Indic} & \textbf{NTX} & \textbf{Nolly} & \textbf{Norw} & \textbf{NusaT} & \textbf{NusaX} & \textbf{Tato} \\
\midrule
Qwen3-4B & \textbf{79.4} & 98.9 & \textbf{25.9} & \textbf{71.3} & \textbf{87.1} & \textbf{74.1} & \textbf{82.0} & \textbf{93.7} & \textbf{87.8} & \textbf{63.6} & \textbf{93.1} & \textbf{91.1} & \textbf{86.9} & \textbf{76.3} \\
jina-v4 & 62.4 & 98.5 & 11.3 & 34.0 & 84.8 & 53.5 & 68.0 & 79.3 & 71.7 & 32.1 & \textbf{93.1} & 63.4 & 64.6 & 57.2 \\
\midrule
Qwen3-0.6B (instr.) & 72.2 & 98.4 & 21.2 & 54.2 & 80.9 & 62.8 & 75.8 & 90.4 & 79.5 & 61.3 & 92.7 & \textbf{85.4} & 78.4 & 58.1 \\
Qwen3-0.6B (generic) & 72.2 & 98.4 & 21.2 & 54.2 & 80.9 & 62.8 & 75.8 & 90.4 & 79.5 & 61.3 & 92.7 & \textbf{85.4} & 78.4 & 58.1 \\
jina-v3 & 65.3 & 98.4 & 10.5 & 37.3 & 85.1 & 55.3 & 72.8 & 87.0 & 78.3 & 33.2 & 92.6 & 62.3 & 64.2 & 71.1 \\
snowflake-l-v2 & 64.1 & 98.1 & 10.3 & 42.0 & 80.5 & 54.7 & 71.4 & 84.7 & 77.4 & 33.6 & 92.5 & 62.8 & 58.2 & 66.9 \\
mult.-e5-l-instr. & \textbf{80.1} & \textbf{99.0} & \textbf{22.0} & 55.4 & \textbf{87.3} & \textbf{86.0} & \textbf{78.9} & \textbf{91.1} & \textbf{93.7} & \textbf{80.7} & 93.5 & 85.1 & \textbf{85.3} & \textbf{83.7} \\
\JEmbeddingVFiveTextSmallShort{} & 69.7 & 98.7 & 13.6 & \textbf{78.0} & 84.8 & 58.0 & 74.9 & 88.5 & 76.6 & 38.8 & \textbf{94.1} & 70.1 & 69.2 & 61.1 \\
\midrule
KaLM-mini-v2.5 & 65.0 & 98.5 & \textbf{14.1} & 43.6 & 82.5 & 58.4 & 63.3 & 78.8 & 74.6 & \textbf{51.5} & 92.7 & \textbf{73.0} & 65.2 & 49.2 \\
voyage-nano & 64.1 & 97.9 & 11.7 & 35.7 & 74.8 & \textbf{60.9} & \textbf{76.3} & \textbf{87.3} & \textbf{77.7} & 38.3 & 93.3 & 61.3 & 53.9 & \textbf{64.6} \\
Gemma-300m & 64.4 & \textbf{98.7} & 12.7 & 34.6 & \textbf{83.9} & 55.3 & 74.4 & 87.1 & 73.9 & 41.3 & 90.8 & 66.1 & \textbf{67.1} & 51.4 \\
\JEmbeddingVFiveTextNanoShort{} & \textbf{67.7} & 98.6 & 12.5 & \textbf{77.1} & \textbf{83.9} & 53.4 & 71.7 & 86.2 & 72.8 & 37.1 & \textbf{94.8} & 69.6 & 65.8 & 56.5 \\
\midrule
v5-small stage 1 & 69.1 & 98.8 & 15.0 & 57.1 & 82.1 & 60.7 & 76.5 & 89.5 & 78.4 & 39.1 & 94.3 & 74.8 & 70.5 & 60.9 \\
v5-nano stage 1 & 67.3 & 98.7 & 14.4 & 56.9 & 81.0 & 57.0 & 74.2 & 87.4 & 75.7 & 38.7 & 93.3 & 72.1 & 69.0 & 56.7 \\
\bottomrule
\end{tabular}

\vspace{2pt}
\begin{flushleft}
\footnotesize
\textbf{Tasks}: Avg:~Average over all tasks,
BUCC:~BUCC.v2,
Bible:~BibleNLP,
Bornh:~Bornholm,
DiaBl:~DiaBla,
Flores:~FLORES,
IN22:~IN22 General,
Indic:~IndicGenBench FLORES,
NTX:~NTREX;
Nolly:~NollySenti,
Norw:~Norwegian Courts,
NusaT:~NusaTranslation,
NusaX:~NusaX,
Tato:~Tatoeba.
\end{flushleft}
\end{table}

\begin{table}[H]
\caption{Evaluation Results on MMTEB Classification Tasks (Accuracy [\%]).}
\label{tab:mteb_multilingual_classification_split}
\centering
\small
\setlength{\tabcolsep}{1.2pt}
\begin{tabular}{l*{22}{c}}
\toprule
\textbf{Model} &
\textbf{Afr} & \textbf{ACF} & \textbf{BgS} & \textbf{CSF} & \textbf{Cat} & \textbf{Cyr} &
\textbf{CzP} & \textbf{DBP} & \textbf{Dal} & \textbf{Est} & \textbf{Fil} &
\textbf{Fin} & \textbf{Grk} & \textbf{Guj} & \textbf{Ind} & \textbf{Idn} &
\textbf{Zul} & \textbf{Ita} & \textbf{Kor} & \textbf{Kur} & \textbf{Mac} & \textbf{Mas} \\
\midrule
Qwen3-4B & \textbf{50.9} & \textbf{92.5} & \textbf{74.9} & \textbf{58.7} & \textbf{52.6} & \textbf{95.0} & \textbf{68.3} & \textbf{99.3} & \textbf{50.7} & \textbf{59.6} & \textbf{47.8} & \textbf{93.9} & \textbf{51.9} & \textbf{92.0} & \textbf{95.1} & \textbf{64.3} & \textbf{23.2} & \textbf{66.7} & \textbf{61.1} & \textbf{81.9} & \textbf{75.5} & \textbf{84.3} \\
jina-v4 & 40.9 & 72.0 & 61.1 & 31.6 & 49.5 & 34.2 & 52.3 & 81.0 & 49.9 & 39.8 & 30.7 & 78.2 & 26.0 & 82.7 & 20.2 & 60.0 & 22.5 & 58.7 & 57.0 & 65.7 & 53.5 & 72.4 \\
\midrule
Qwen3-0.6B (instr.) & \textbf{46.0} & 90.4 & 73.4 & 44.1 & 49.1 & 90.2 & \textbf{63.8} & \textbf{98.8} & 50.0 & 41.1 & 40.8 & \textbf{90.6} & 39.7 & 85.8 & 93.1 & \textbf{64.8} & 25.6 & 61.7 & 55.9 & 80.1 & 61.0 & 80.5 \\
Qwen3-0.6B (gen.) & 40.0 & 70.1 & 65.4 & 32.3 & 47.9 & 74.3 & 50.9 & 96.8 & 50.0 & 34.6 & 28.6 & 74.6 & 39.3 & 81.3 & 36.5 & 59.0 & 23.6 & 70.1 & 58.1 & 71.1 & 47.2 & 77.0 \\
jina-v3 & 42.8 & 92.2 & 76.0 & 39.1 & 45.1 & 43.3 & 61.9 & 76.1 & 50.1 & 48.8 & 38.5 & 79.5 & 11.6 & 86.2 & 18.0 & 58.1 & 20.8 & 62.8 & 56.2 & 57.6 & 61.2 & 71.5 \\
snowflake-l-v2 & 44.1 & 64.7 & 59.3 & 31.0 & 49.1 & 57.7 & 49.7 & 89.6 & \textbf{50.2} & 40.7 & 31.8 & 72.0 & 35.7 & 84.4 & 33.0 & 60.9 & 23.9 & 68.1 & \textbf{58.4} & 61.7 & 61.3 & 74.3 \\
mult.-e5-l-instr. & 45.4 & 68.6 & \textbf{78.6} & 43.0 & 51.0 & 81.1 & 60.7 & 95.5 & 50.0 & 49.1 & 40.8 & 84.4 & 32.3 & 87.5 & 86.3 & 61.3 & \textbf{37.3} & 62.9 & 58.2 & 80.9 & 66.7 & 80.5 \\
\JEmbeddingVFiveTextSmallShort{} & 36.7 & \textbf{94.3} & 74.8 & \textbf{46.7} & \textbf{65.9} & \textbf{97.6} & 63.6 & 98.3 & 49.9 & \textbf{56.1} & \textbf{41.9} & 63.4 & \textbf{65.0} & \textbf{93.0} & \textbf{98.0} & 57.7 & 19.9 & \textbf{91.0} & 55.2 & \textbf{93.4} & \textbf{67.7} & \textbf{86.9} \\
\midrule
KaLM-mini-v2.5 & 42.6 & \textbf{95.5} & 63.3 & 33.1 & 50.1 & 80.7 & 52.0 & 95.3 & 50.2 & 37.9 & 32.6 & 84.0 & 28.1 & 74.0 & 75.5 & 59.0 & 26.9 & 72.0 & 57.6 & 60.0 & 52.3 & 76.4 \\
voyage-nano & 42.7 & 66.1 & 66.2 & \textbf{36.3} & 51.1 & 39.2 & 52.4 & 90.5 & 49.9 & 40.0 & 31.4 & 79.3 & 38.3 & 84.9 & 18.7 & 57.6 & \textbf{29.3} & 66.1 & 57.2 & 74.9 & 53.9 & 77.2 \\
Gemma-300M & \textbf{44.5} & 84.2 & 71.3 & 34.5 & 51.2 & 58.6 & 58.6 & 94.3 & 50.3 & 38.3 & \textbf{40.5} & \textbf{86.4} & 29.0 & 82.8 & 46.6 & \textbf{60.9} & 26.4 & 70.4 & \textbf{58.1} & 60.0 & 45.3 & 74.9 \\
\JEmbeddingVFiveTextNanoShort{} & 36.8 & 93.7 & \textbf{74.0} & 40.6 & \textbf{65.8} & \textbf{97.9} & \textbf{62.7} & \textbf{97.8} & \textbf{50.4} & \textbf{51.0} & \textbf{40.5} & 48.6 & \textbf{59.0} & \textbf{91.0} & \textbf{97.6} & 54.3 & 17.8 & \textbf{87.4} & 53.7 & \textbf{93.3} & \textbf{64.9} & \textbf{82.7} \\
\midrule
v5-small stage 1 & 42.2 & 75.5 & 56.1 & 31.8 & 49.1 & 65.2 & 52.0 & 91.4 & 50.1 & 35.9 & 30.0 & 77.8 & 45.8 & 78.4 & 41.4 & 59.0 & 26.3 & 71.5 & 56.5 & 67.8 & 49.7 & 76.3 \\
v5-nano stage 1 & 41.9 & 72.7 & 54.7 & 30.7 & 49.4 & 72.9 & 50.8 & 91.2 & 49.9 & 36.8 & 28.4 & 77.1 & 44.1 & 76.1 & 57.2 & 58.9 & 25.6 & 67.5 & 56.1 & 67.9 & 50.4 & 72.9 \\
\midrule
\textbf{Model} &
\textbf{MI} & \textbf{MH} & \textbf{Nep} & \textbf{Nor} & \textbf{NE} & \textbf{NX} &
\textbf{Odi} & \textbf{PAC} & \textbf{Poe} & \textbf{PE} & \textbf{Pun} &
\textbf{Sca} & \textbf{Hin} & \textbf{Sin} & \textbf{Sis} & \textbf{Slk} &
\textbf{Swa} & \textbf{CH} & \textbf{Tox} & \textbf{Tsw} & \textbf{TT} & \textbf{Avg} \\
\midrule
Qwen3-4B & 76.5 & 77.5 & 97.3 & 90.8 & 60.4 & 79.7 & 93.8 & 69.7 & 72.2 & 77.1 & 81.9 & 51.3 & 76.4 & 75.5 & 57.3 & 93.3 & 67.2 & 62.6 & 91.4 & 36.4 & 81.7 & 72.3 \\
jina-v4 & 70.5 & 59.5 & 93.4 & 46.5 & 43.3 & 63.7 & 70.1 & 64.3 & 55.3 & 24.6 & 80.5 & 50.0 & 54.2 & 58.8 & 51.1 & 68.9 & 58.9 & 55.3 & 64.5 & 29.5 & 71.9 & 55.2 \\
\midrule
Qwen3-0.6B (instr.) & 61.4 & 64.3 & 95.6 & 84.3 & 47.8 & 71.3 & 84.2 & 69.5 & 73.3 & 71.5 & 83.5 & 50.5 & \textbf{74.3} & 59.6 & \textbf{58.0} & 86.3 & 58.4 & 55.6 & 82.1 & 34.9 & 80.8 & 66.8 \\
Qwen3-0.6B (gen.) & 50.0 & 56.6 & 95.0 & 65.1 & 41.7 & 63.1 & 78.6 & 67.4 & 56.9 & 35.1 & 82.4 & 50.4 & 56.5 & 57.7 & 57.0 & 72.6 & \textbf{61.6} & 54.8 & 68.0 & 35.4 & 77.9 & 58.4 \\
jina-v3 & 68.5 & 60.3 & 93.4 & 40.9 & 42.2 & 66.0 & 81.4 & 69.2 & 59.1 & 59.7 & 77.5 & 50.2 & 65.7 & 74.3 & 51.4 & 84.3 & 56.2 & 55.1 & 91.3 & 25.9 & 57.1 & 58.8 \\
snowflake-l-v2 & 63.0 & 59.1 & 91.2 & 49.1 & 42.7 & 65.5 & 82.9 & 66.9 & 51.2 & 34.5 & 81.1 & \textbf{50.6} & 56.0 & 73.3 & 57.9 & 66.2 & 60.9 & 57.1 & 65.9 & 25.4 & 65.6 & 57.4 \\
mult.-e5-l-instr. & 62.7 & \textbf{64.7} & 97.0 & 76.6 & 44.1 & 71.0 & 84.5 & 65.7 & 57.1 & 57.0 & 84.4 & 50.4 & 59.6 & \textbf{78.0} & 47.5 & 87.5 & 59.1 & 55.5 & 66.8 & 46.4 & 74.9 & 64.9 \\
\JEmbeddingVFiveTextSmallShort{} & \textbf{85.3} & 58.3 & \textbf{99.5} & \textbf{88.1} & \textbf{75.8} & \textbf{79.7} & \textbf{95.8} & \textbf{87.0} & \textbf{81.9} & \textbf{73.9} & \textbf{88.6} & 50.2 & 44.3 & 52.5 & 46.6 & \textbf{87.7} & 54.6 & \textbf{66.1} & \textbf{93.7} & \textbf{53.8} & \textbf{86.3} & \textbf{71.3} \\
\midrule
KaLM-mini-v2.5 & 83.2 & \textbf{62.6} & 95.3 & 62.1 & 39.2 & 65.3 & 64.8 & 65.1 & 57.7 & 42.4 & 76.7 & 50.1 & 55.7 & 60.3 & 55.1 & 72.8 & 57.7 & 53.7 & 91.7 & 42.5 & 76.9 & 61.2 \\
voyage-nano & 57.9 & 58.3 & 94.9 & 42.5 & 44.7 & 65.5 & 84.7 & 65.9 & 52.3 & 47.9 & 81.1 & 50.2 & 54.6 & \textbf{76.2} & \textbf{64.4} & 73.2 & 63.1 & 59.3 & 64.7 & 41.5 & 72.4 & 58.6 \\
Gemma-300M & 62.7 & 61.0 & 95.5 & 65.6 & 44.2 & 69.8 & 57.9 & 67.9 & 58.9 & 62.8 & \textbf{82.4} & \textbf{50.8} & \textbf{65.5} & 65.7 & 57.0 & 73.3 & \textbf{66.0} & 57.7 & 82.9 & 31.1 & 73.0 & 60.9 \\
\JEmbeddingVFiveTextNanoShort{} & \textbf{84.1} & 57.0 & \textbf{98.6} & \textbf{88.2} & \textbf{73.4} & \textbf{76.1} & \textbf{94.5} & \textbf{86.3} & \textbf{75.6} & \textbf{66.9} & 80.4 & 50.1 & 55.4 & 51.7 & 48.5 & \textbf{85.1} & 50.7 & \textbf{60.5} & \textbf{92.8} & \textbf{52.6} & \textbf{84.7} & \textbf{69.2} \\
\midrule
v5-small stage 1 & 56.7 & 58.2 & 92.7 & 61.2 & 45.0 & 65.0 & 74.9 & 64.4 & 54.2 & 32.4 & 80.1 & 50.1 & 55.9 & 66.3 & 61.5 & 65.8 & 63.1 & 55.9 & 69.9 & 35.3 & 70.9 & 58.4 \\
v5-nano stage 1 & 56.8 & 57.7 & 93.8 & 65.4 & 44.2 & 65.3 & 72.6 & 64.5 & 54.1 & 36.1 & 68.7 & 50.0 & 56.6 & 62.2 & 59.4 & 65.3 & 61.3 & 56.4 & 68.1 & 32.7 & 72.5 & 58.1 \\
\bottomrule
\end{tabular}

\vspace{2pt}
\begin{flushleft}
\footnotesize
\textbf{Tasks}:
Avg:~Average over all tasks,
Afr:~AfriSentiClassification,
ACF:~AmazonCounterfactualClassification,
BgS:~BulgarianStoreReviewSentimentClassification,
CSF:~CzechCSFDClassification,
Cat:~CataloniaTweetsClassification,
Cyr:~CyprusTurkishTweetsSentiment,
CzP:~CzechProductReviewsClassification,
DBP:~DBPediaClassification,
Dal:~DalajClassification,
Est:~EstonianValenceClassification,
Fil:~FilipinoHateSpeechClassification,
Fin:~FinnishClassification,
Grk:~GreekLegalClassification,
Guj:~GujaratiNewsClassification,
Ind:~IndicSentimentClassification,
Idn:~IndonesianClassification,
Zul:~IsiZuluSentimentClassification,
Ita:~ItalianClassification,
Kor:~KoreanSarcasmClassification,
Kur:~KurdishSentimentClassification,
Mac:~MacedonianClassification,
Mas:~MasakhaNEWSClassification,
MI:~MassiveIntentClassification,
MH:~MultilingualHateSpeechClassification,
Nep:~NepaliClassification,
Nor:~NordicSentimentClassification,
NE:~NusaXEmotionClassification,
NX:~NusaXClassification,
Odi:~OdiaNewsClassification,
PAC:~PAWSXClassification,
Poe:~PoemSentimentClassification,
PE:~PolishEmotionClassification,
Pun:~PunjabiClassification,
Sca:~ScalaSentimentClassification,
Hin:~HindiClassification,
Sin:~SinhalaClassification,
Sis:~SiswatiClassification,
Slk:~SlovakClassification,
Swa:~SwahiliClassification,
CH:~SwissJudgementClassification,
Tox:~ToxicConversationsClassification,
Tsw:~XitsongaClassification,
TT:~TwitterTopicClassification
\end{flushleft}
\end{table}

\begin{table}[H]
\caption{Evaluation Results on MMTEB Multi-Label Classification Tasks (Accuracy [\%]).}
\label{tab:mteb_multilingual_multi_label_classification}
\centering
\small
\setlength{\tabcolsep}{3pt}
\begin{tabular}{l*{6}{c}}
\toprule
\textbf{Model} & \textbf{Avg.} & \textbf{BrazilianToxic} & \textbf{CEDR} & \textbf{KorHate} & \textbf{MalteseNews} & \textbf{MultiEURLEX} \\
\midrule
Qwen3-4B & \textbf{26.8} & \textbf{20.6} & \textbf{51.0} & \textbf{14.9} & \textbf{42.1} & \textbf{5.1} \\
jina-v4 & 19.3 & 17.6 & 40.8 & 7.8 & 26.7 & 3.8 \\
\midrule
Qwen3-0.6B (instr.) & 24.6 & 22.6 & 49.9 & 9.7 & 36.1 & 4.6 \\
Qwen3-0.6B (generic) & 21.1 & 23.7 & 38.7 & 8.8 & 29.7 & 4.4 \\
jina-v3 & 18.4 & 19.7 & 47.4 & 10.6 & 12.7 & 1.5 \\
snowflake-l-v2 & 18.9 & 22.5 & 38.5 & 10.5 & 18.7 & 4.6 \\
mult.-e5-l-instr. & 22.9 & 19.8 & 50.0 & 10.2 & 29.0 & 5.5 \\
\JEmbeddingVFiveTextSmallShort{} & \textbf{42.0} & \textbf{21.3} & \textbf{65.0} & \textbf{60.1} & \textbf{57.0} & \textbf{6.6} \\
\midrule
KaLM-mini-v2.5 & 21.0 & \textbf{22.9} & 40.6 & 7.6 & 29.2 & 4.6 \\
voyage-nano & 20.1 & 17.3 & 41.6 & 8.8 & 29.1 & 3.6 \\
Gemma-300M & 24.8 & 22.3 & 52.8 & 11.6 & 33.1 & 4.3 \\
\JEmbeddingVFiveTextNanoShort{} & \textbf{41.3} & 20.7 & \textbf{65.2} & \textbf{58.8} & \textbf{56.2} & \textbf{5.5} \\
\midrule
v5-small stage 1 & 20.3 & 18.5 & 41.4 & 8.9 & 28.1 & 4.5 \\
v5-nano stage 1 & 19.9 & 18.2 & 40.9 & 8.5 & 27.7 & 4.3 \\
\bottomrule
\end{tabular}
\begin{flushleft}
\textbf{Tasks}: Avg:~Average over all tasks, BrazilianToxic:~BrazilianToxicTweetsClassification, 
CEDR:~CEDRClassification, 
KorHate:~KorHateSpeechMLClassification, 
MalteseNews:~MalteseNewsClassification, 
MultiEURLEX:~MultiEURLEXMultilabelClassification.
\end{flushleft}
\end{table}

\begin{table}[H]
\caption{Evaluation Results on MMTEB Clustering Tasks (V-measure [\%] - see footnote in Section \ref{eval:mteb})}
\label{tab:mteb_multilingual_v2_clustering}
\centering
\small
\setlength{\tabcolsep}{1.5pt}
\begin{tabular}{l*{17}{c}}
\toprule
\textbf{Model} & \textbf{Avg} & \textbf{Allo} & \textbf{AXP} & \textbf{AXS} & \textbf{BigPat} & \textbf{BioP} & \textbf{CLSP} & \textbf{HalS} & \textbf{MNC} & \textbf{MedP} & \textbf{PlscP} & \textbf{Rom} & \textbf{S200} & \textbf{SEx} & \textbf{SCP} & \textbf{Cities} & \textbf{WikiP} \\
\midrule
Qwen3-4B & \textbf{57.2} & \textbf{59.5} & \textbf{64.8} & \textbf{65.2} & \textbf{43.2} & \textbf{50.9} & \textbf{73.2} & \textbf{30.5} & \textbf{56.2} & \textbf{45.3} & \textbf{75.1} & \textbf{44.3} & \textbf{41.3} & \textbf{77.6} & \textbf{62.1} & \textbf{93.6} & \textbf{31.6} \\
jina-v4 & 44.0 & 44.5 & 57.9 & 57.6 & 28.7 & 38.4 & 37.4 & 25.1 & 40.5 & 35.2 & 69.6 & 41.1 & 27.4 & 54.0 & 27.3 & 92.0 & 26.8 \\
\midrule
Qwen3-0.6B (instr.) & 52.3 & 54.0 & 63.7 & \textbf{63.8} & 32.5 & 47.3 & \textbf{62.0} & 29.0 & 53.2 & 42.2 & 74.2 & 40.3 & 34.1 & \textbf{71.2} & 53.4 & 86.8 & 29.7 \\
Qwen3-0.6B (generic) & 49.8 & 53.8 & 63.2 & 61.8 & 32.3 & 45.1 & 48.0 & 28.8 & 50.4 & 40.7 & 72.8 & 39.7 & 34.8 & 66.2 & 49.8 & 80.9 & 28.6 \\
jina-v3 & 45.7 & 44.7 & 58.9 & 55.9 & 37.1 & 40.0 & 39.4 & 29.3 & 46.2 & 38.2 & 71.5 & 40.6 & 32.0 & 56.7 & 26.8 & 85.3 & 27.8 \\
snowflake-l-v2 & 42.8 & 45.7 & 57.2 & 53.1 & 34.1 & 37.2 & 34.2 & 24.9 & 42.9 & 35.0 & 71.1 & 39.6 & 25.4 & 56.3 & 26.0 & 74.2 & 27.1 \\
mult.-e5-l-instr. & 50.8 & \textbf{56.5} & 62.5 & 61.3 & \textbf{43.2} & 42.7 & 42.4 & 30.1 & 59.2 & 38.1 & 72.7 & \textbf{40.9} & \textbf{47.2} & 60.0 & 47.6 & 76.2 & \textbf{31.5} \\
\JEmbeddingVFiveTextSmallShort{} & \textbf{53.4} & 53.0 & \textbf{65.8} & 62.7 & 42.6 & \textbf{47.7} & 50.7 & \textbf{31.7} & \textbf{53.2} & \textbf{43.5} & \textbf{74.4} & 40.4 & 40.5 & 70.1 & \textbf{57.6} & \textbf{89.8} & 30.8 \\
\midrule
KaLM-mini-v2.5 & \textbf{53.8} & 52.0 & 63.5 & \textbf{61.2} & 43.3 & 50.7 & \textbf{68.3} & 29.2 & \textbf{54.1} & \textbf{45.6} & 73.9 & 40.3 & 37.5 & 75.8 & 49.9 & 85.8 & \textbf{30.5} \\
voyage-nano & 45.4 & 49.0 & 58.2 & 56.4 & 31.0 & 41.2 & 38.5 & 26.7 & 42.0 & 38.3 & 71.5 & \textbf{42.1} & 32.8 & 52.3 & 33.6 & 86.0 & 26.9 \\
Gemma-300m & 51.2 & 52.8 & 63.6 & 59.6 & 41.6 & \textbf{52.1} & 41.5 & \textbf{29.3} & 43.5 & 44.1 & 72.1 & 41.9 & 26.5 & \textbf{90.9} & 40.0 & \textbf{92.0} & 27.0 \\
\JEmbeddingVFiveTextNanoShort{} & 52.7 & \textbf{56.4} & \textbf{65.0} & 61.1 & \textbf{43.5} & 46.4 & 48.9 & 29.2 & 52.3 & 42.6 & \textbf{74.4} & 40.6 & \textbf{38.9} & 69.4 & \textbf{56.0} & 88.7 & 30.2 \\
\midrule
v5-small stage 1 & 44.7 & 41.1 & 58.2 & 57.0 & 32.5 & 41.8 & 40.4 & 24.4 & 37.2 & 37.8 & 71.2 & 42.5 & 27.7 & 55.4 & 31.7 & 88.8 & 27.5 \\
v5-nano stage 1 & 45.6 & 45.5 & 58.4 & 57.7 & 33.8 & 41.2 & 39.9 & 25.2 & 40.1 & 38.1 & 72.2 & 43.1 & 28.7 & 58.4 & 33.4 & 85.6 & 27.5 \\

\bottomrule
\end{tabular}
\begin{flushleft}
\textbf{Tasks}: Avg:~Average over all tasks, 
Allo:~AlloProfClusteringS2S.v2,
AXP:~ArXivHierarchicalClusteringP2P,
AXS:~ArXivHierarchicalClusteringS2S,
BigPat:~BigPatentClustering.v2,
BioP:~BiorxivClusteringP2P.v2,
CLSP:~CLSClusteringP2P.v2,
HalS:~HALClusteringS2S.v2,
MNC:~MasakhaNEWSClusteringS2S,
MedP:~MedrxivClusteringP2P.v2,
PlscP:~PlscClusteringP2P.v2,
Rom:~RomaniBibleClustering,
S200:~SIB200ClusteringS2S,
SEx:~StackExchangeClustering.v2,
SCP:~SwednClusteringP2P,
Cities:~WikiCitiesClustering,
WikiP:~WikiClusteringP2P.v2.
\end{flushleft}
\end{table}

\begin{table}[H]
\caption{Evaluation Results on Instruction Reranking Tasks (p-MRR [\%]).}
\label{tab:instruction_reranking}
\centering
\small
\setlength{\tabcolsep}{4pt}
\begin{tabular}{l*{4}{c}}
\toprule
\textbf{Model} & \textbf{Avg} & \textbf{Core17} & \textbf{News21} & \textbf{Robust04} \\
\midrule
Qwen3-4B & \textbf{11.56} & \textbf{13.53} & \textbf{8.71} & \textbf{12.44} \\
jina-v4 & 0.71 & 3.24 & 0.34 & $-$1.46 \\
\midrule
Qwen3-0.6B (instr.) & \textbf{5.09} & \textbf{8.93} & \textbf{3.61} & \textbf{2.75} \\
Qwen3-0.6B (generic) & 3.84 & 6.30 & 2.59 & 2.63 \\
jina-v3 & $-$1.34 & $-$0.06 & 2.36 & $-$6.31 \\
snowflake-l-v2 & $-$2.47 & 0.43 & $-$0.24 & $-$7.60 \\
mult.-e5-l-instr. & $-$0.40 & 1.82 & 1.50 & $-$4.52 \\
\JEmbeddingVFiveTextSmallShort{} & 1.34 & 2.40 & 3.24 & $-$1.61 \\
\midrule
KaLM-mini-v2.5 & $-$0.56 & 0.66 & 0.46 & $-$2.80 \\
voyage-4-nano & \textbf{5.61} & \textbf{6.32} & \textbf{11.45} & $-$0.94 \\
Gemma-300M & 3.49 & 5.13 & 4.60 & \textbf{0.73} \\
\JEmbeddingVFiveTextNanoShort{} & 0.05 & 1.78 & 1.28 & $-$2.92 \\
\midrule
v5-small stage 1 & $-$0.32 & 1.58 & 0.00 & $-$2.53 \\
v5-nano stage 1 & $-$1.31 & 0.63 & 0.18 & $-$4.75 \\
\bottomrule
\end{tabular}
\begin{flushleft}
\textbf{Tasks}: Avg:~Average over all tasks, Core17:~Core17InstructionRetrieval, News21:~News21InstructionRetrieval, Robust04:~Robust04InstructionRetrieval.
\end{flushleft}
\end{table}

\subsection{Other Retrieval Benchmark}
\label{app:retrieval-benchmarks}

\begin{table}[H]
\caption{Retrieval performance on BeIR (nDCG@10 [\%]).}
\label{tab:beir_retrieval}
\centering
\small
\setlength{\tabcolsep}{1.5pt}
\begin{tabular}{l*{16}{c}}
\toprule
\textbf{Model} & \textbf{Avg} & \textbf{Arg} & \textbf{CQA} & \textbf{CF} & \textbf{DB} & \textbf{FEV} & \textbf{FiQA} & \textbf{HPQA} & \textbf{MSM} & \textbf{NFC} & \textbf{NQ} & \textbf{Quora} & \textbf{SCD} & \textbf{SCF} & \textbf{TREC} & \textbf{TOU} \\
\midrule
Qwen3-4B & \textbf{61.6} & \textbf{75.6} & \textbf{50.3} & \textbf{47.4} & \textbf{48.2} & \textbf{91.6} & \textbf{62.7} & \textbf{74.7} & \textbf{42.7} & \textbf{41.1} & \textbf{63.1} & \textbf{88.1} & \textbf{31.4} & \textbf{78.3} & \textbf{92.9} & \textbf{35.4} \\
jina-v4 & 54.0 & 67.0 & 43.7 & 35.1 & 43.9 & 87.9 & 48.5 & 68.5 & 38.1 & 34.4 & 61.7 & 78.6 & 21.5 & 76.1 & 80.4 & 24.1 \\
\midrule
Qwen3-0.6B & 55.5 & \textbf{71.0} & 46.0 & \textbf{42.4} & 39.5 & 88.1 & 46.6 & 65.7 & 38.0 & 36.7 & 53.5 & 87.8 & \textbf{24.4} & 69.7 & \textbf{90.5} & \textbf{33.2} \\
jina-v3 & 53.2 & 43.3 & 42.6 & \textbf{42.4} & 41.0 & 89.1 & 47.4 & 64.7 & 40.8 & 36.6 & \textbf{64.2} & \textbf{89.2} & 19.9 & 72.5 & 77.7 & 26.3 \\
snowflake-l-v2 & 55.2 & 59.1 & 45.9 & 41.8 & 43.4 & \textbf{91.5} & 45.4 & 68.2 & \textbf{44.9} & 35.1 & 63.7 & 88.8 & 20.3 & 70.9 & 83.6 & 25.9 \\
mult.-e5-l-instr. & 52.7 & 58.5 & 44.3 & 29.9 & 38.4 & 78.0 & 48.4 & \textbf{69.3} & 40.4 & 36.3 & 57.8 & \textbf{89.2} & 19.2 & 71.6 & 82.5 & 27.4 \\
\JEmbeddingVFiveTextSmallShort{} & \textbf{56.7} & 65.1 & \textbf{46.7} & 41.5 & \textbf{44.4} & 90.0 & \textbf{49.6} & 69.8 & 42.1 & \textbf{39.8} & 64.0 & 89.1 & 23.0 & \textbf{76.5} & 78.5 & 29.9 \\
\midrule
KaLM-mini-v2.5 & 55.0 & 60.2 & \textbf{47.2} & 34.5 & 42.6 & 87.9 & 47.1 & \textbf{71.8} & 40.6 & 37.1 & 58.6 & \textbf{89.6} & 21.6 & 74.4 & \textbf{83.0} & 28.9 \\
voyage-4-nano & 49.9 & 58.6 & 44.3 & 22.2 & 39.5 & 68.4 & \textbf{51.0} & 62.0 & 31.5 & \textbf{39.6} & 49.2 & 86.1 & 21.3 & 75.2 & 77.9 & 22.2 \\
Gemma-300M & 53.7 & \textbf{66.0} & 42.1 & 26.7 & 44.4 & 81.1 & 47.4 & 70.1 & 38.6 & 39.2 & \textbf{63.5} & 86.6 & 19.5 & \textbf{78.7} & 76.4 & 25.1 \\
\JEmbeddingVFiveTextNanoShort{} & \textbf{56.1} & 65.7 & 44.6 & \textbf{39.6} & \textbf{45.3} & \textbf{89.5} & 47.9 & 69.1 & \textbf{41.6} & 38.7 & 63.4 & 88.9 & \textbf{22.6} & 75.8 & 77.6 & \textbf{30.7} \\
\midrule
v5-small stage 1 & 54.9 & 64.0 & 46.0 & 38.0 & 43.1 & 86.2 & 47.5 & 68.0 & 38.9 & 38.3 & 57.3 & 88.6 & 22.3 & 74.5 & 79.5 & 30.6 \\
v5-nano stage 1 & 54.9 & 65.6 & 44.5 & 36.3 & 43.6 & 87.3 & 46.2 & 68.1 & 40.5 & 38.0 & 57.1 & 88.1 & 22.2 & 73.1 & 79.4 & 33.8 \\
\bottomrule
\end{tabular}
\begin{flushleft}
    \textbf{Tasks}:  Avg:~Average over all tasks, Arg:~ArguAna, CQA:~CQADupstackRetrieval, CF:~ClimateFEVER, DB:~DBPedia, FEV:~FEVER, FiQA:~FiQA2018, HPQA:~HotpotQA, MSM:~MSMARCO, NFC:~NFCorpus, NQ:~Natural~Questions, Quora:~QuoraRetrieval, SCD:~SCIDOCS, SCF:~SciFact, TREC:~TRECCOVID, TOU:~Touche2020
\end{flushleft}
\end{table}

\begin{table}[H]
\caption{Retrieval performance on LongEmbed (nDCG@10 [\%])}
\label{tab:longembed}
\centering
\small
\begin{tabular}{@{}lcccccccc@{}}
\toprule
\textbf{Model} & \textbf{Avg} & \textbf{NaQA} & \textbf{Needle*} & \textbf{Passkey*} & \textbf{QMSum} & \textbf{SummScreen} & \textbf{Wikim} \\ 
\midrule
Qwen3-4B & \textbf{78.82} & \textbf{68.94} & \textbf{75.50} & \textbf{84.25} & \textbf{52.35} & \textbf{97.96} & \textbf{93.92} \\
jina-v4 & 69.88 & 58.71 & 60.75 & 69.75 & 46.06 & 96.96 & 87.02 \\
\midrule
Qwen3-0.6B & \textbf{72.20} & \textbf{63.25} & 50.75 & \textbf{84.75} & \textbf{47.70} & 96.72 & \textbf{90.00} \\
jina-v3 & 55.67 & 34.30 & \textbf{64.00} & 38.00 & 39.34 & 92.33 & 66.02 \\
snowflake-l-v2 & 63.74 & 43.63 & 50.25 & 77.25 & 40.08 & 96.38 & 74.84 \\
mult.-e5-l-instr. & 41.76 & 26.71 & 29.50 & 37.75 & 26.08 & 72.75 & 57.79 \\
\JEmbeddingVFiveTextSmallShort{} & 66.39 & 52.95 & 44.50 & 80.50 & 43.80 & \textbf{96.88} & 79.68 \\
\midrule
KaLM-mini-v2.5 & 43.35 & 29.32 & 31.50 & 38.25 & 27.06 & 74.38 & 59.61 \\
voyage-4-nano & \textbf{74.93} & \textbf{63.71} & \textbf{61.00} & \textbf{87.25} & \textbf{51.44} & \textbf{98.46} & \textbf{87.72} \\
Gemma-300M & 55.29 & 28.83 & 41.25 & 61.00 & 37.62 & 91.19 & 71.88 \\
\JEmbeddingVFiveTextNanoShort{} & 63.65 & 52.17 & 59.75 & 81.50 & 31.83 & 81.80 & 74.87 \\
\midrule
v5-small stage 1 & 68.36 & 56.85 & 46.00 & 83.00 & 46.39 & 97.16 & 80.76 \\
v5-nano stage 1 & 63.48 & 46.17 & 64.00 & 83.00 & 34.40 & 82.73 & 70.57 \\
v5-small pre-long-ctx** & 44.54 & 18.36 & 38.25 & 47.25 & 27.78 & 74.26 & 61.33 \\
\bottomrule
\end{tabular}
\medskip
\begin{flushleft}
\textbf{Tasks:} Avg:~Average over all tasks, NaQA:~LEMBNarrativeQARetrieval, Needle:~LEMBNeedleRetrieval, Passkey:~LEMBPasskeyRetrieval, QMSum:~LEMBQMSumRetrieval, SummScreen:~LEMBSummScreenFDRetrieval, Wikim:~LEMBWikimQARetrieval \\
* Scores are in nDCG@1 \quad
** 1st stage checkpoint before long context training was applied 
\end{flushleft}
\end{table}

\begin{table}[H]
\caption{Retrieval performance on RTEB (Public) (nDCG@10 [\%]).}
\label{tab:rteb_public_retrieval}
\centering
\small
\setlength{\tabcolsep}{1.2pt}
\begin{tabular}{l*{17}{c}}
\toprule
\textbf{Model} &
\textbf{Avg} &
\textbf{ACD} &
\textbf{AST} &
\textbf{LS} &
\textbf{LQA} &
\textbf{FinB} &
\textbf{HC3} &
\textbf{FQA} &
\textbf{HuE} &
\textbf{MBPP} &
\textbf{MIR} &
\textbf{Apps} &
\textbf{DS1K} &
\textbf{WSQL} &
\textbf{CDR} &
\textbf{CURE} &
\textbf{Fsh} \\
\midrule
Qwen3-4B & \textbf{70.8} & 39.4 & \textbf{81.2} & \textbf{66.4} & \textbf{66.7} & 77.5 & \textbf{68.9} & 63.4 & \textbf{98.4} & \textbf{91.4} & \textbf{69.5} & \textbf{89.2} & \textbf{64.1} & 84.7 & \textbf{71.5} & \textbf{56.8} & \textbf{43.2} \\
jina-v4 & 66.5 & \textbf{45.2} & 50.1 & 59.9 & 63.9 & \textbf{79.9} & 63.4 & \textbf{68.4} & 97.2 & 89.9 & 62.9 & 78.3 & \textbf{64.1} & \textbf{96.1} & 64.1 & 53.2 & 27.5 \\
\midrule
Qwen3-0.6B & 64.2 & 36.1 & \textbf{79.0} & 63.6 & 53.4 & 74.8 & 54.5 & 56.3 & 92.3 & 87.0 & 61.2 & \textbf{75.3} & 59.7 & 86.8 & 62.5 & 47.0 & 37.9 \\
jina-v3 & 54.6 & 34.8 & 32.8 & 59.2 & 59.4 & 72.2 & \textbf{61.5} & 39.2 & 80.6 & 83.2 & 62.6 & 29.0 & 50.1 & 68.0 & 64.2 & 46.3 & 30.5 \\
snowflake-l-v2 & 54.0 & 34.0 & 22.8 & 66.2 & \textbf{65.5} & 75.6 & 54.4 & \textbf{56.6} & 71.5 & 80.2 & \textbf{66.5} & 9.7 & 42.7 & 69.7 & 60.7 & \textbf{54.7} & 32.3 \\
mult.-e5-l-instr. & 54.8 & 33.3 & 29.7 & \textbf{68.1} & 51.2 & 79.7 & 51.2 & 45.1 & 86.3 & 83.6 & 57.7 & 34.9 & 49.4 & 80.7 & 55.2 & 42.8 & 27.6 \\
\JEmbeddingVFiveTextSmallShort{} & \textbf{66.8} & \textbf{43.9} & 53.3 & 64.9 & 63.6 & \textbf{80.6} & 62.1 & 55.7 & \textbf{96.1} & \textbf{90.5} & 66.6 & 73.3 & \textbf{61.4} & \textbf{93.7} & \textbf{71.1} & 53.6 & \textbf{39.2} \\
\midrule
KaLM-mini-v2.5 & 56.5 & 34.4 & 36.5 & 64.4 & 42.9 & 77.7 & 61.9 & 48.1 & 89.7 & 85.0 & 53.8 & 35.9 & 55.9 & 78.6 & 67.2 & 42.6 & 29.6 \\
voyage-4-nano & \textbf{70.4} & \textbf{42.8} & 48.7 & \textbf{72.3} & \textbf{70.2} & \textbf{90.9} & \textbf{63.1} & \textbf{82.5} & 98.8 & \textbf{91.4} & 58.7 & 81.0 & \textbf{63.9} & \textbf{98.0} & 67.0 & 53.5 & \textbf{43.1} \\
Gemma-300M & 63.8 & 32.9 & 30.7 & 68.4 & 60.1 & 77.8 & 59.8 & 54.7 & \textbf{99.0} & 88.4 & 64.6 & \textbf{84.1} & 57.3 & 91.5 & 63.4 & \textbf{57.1} & 30.5 \\
\JEmbeddingVFiveTextNanoShort{} & 64.1 & 39.7 & \textbf{51.5} & 65.8 & 57.8 & 78.8 & 57.8 & 57.8 & 92.1 & 86.7 & \textbf{65.8} & 58.4 & 54.9 & 97.7 & \textbf{69.4} & 52.8 & 38.3 \\
\midrule
v5-small stage 1 & 64.1 & 40.9 & 49.5 & 68.7 & 62.2 & 80.2 & 58.0 & 57.8 & 94.1 & 88.7 & 64.6 & 63.0 & 60.0 & 79.4 & 67.9 & 52.8 & 38.4 \\
v5-nano stage 1 & 61.1 & 39.1 & 47.0 & 66.9 & 55.5 & 76.5 & 54.7 & 56.8 & 90.6 & 85.5 & 63.9 & 45.1 & 51.3 & 88.8 & 67.2 & 51.2 & 38.0 \\
\bottomrule
\end{tabular}
\begin{flushleft}
\textbf{Tasks}:
Avg:~Average over all tasks,
ACD:~AILACasedocs,
AST:~AILAStatutes,
LS:~LegalSummarization,
LQA:~LegalQuAD,
FinB:~FinanceBenchRetrieval,
HC3:~HC3FinanceRetrieval,
FQA:~FinQARetrieval,
HuE:~HumanEvalRetrieval,
MBPP:~MBPPRetrieval,
MIR:~MIRACLRetrievalHardNegatives,
Apps:~AppsRetrieval,
DS1K:~DS1000Retrieval,
WSQL:~WikiSQLRetrieval,
CDR:~ChatDoctorRetrieval,
CURE:~CUREv1,
Fsh:~FreshStackRetrieval
\end{flushleft}
\end{table}

\subsection{Learning Rate Ablation}
\label{app:learning-rate-ablation}

In this section, we detail the experimental setup used to determine the optimal hyperparameters for the training objectives discussed in Section~\ref{sec:ablation:training-obj}. All experiments in this ablation were conducted using the \texttt{S2ORC} dataset with a fixed student-side trainable projection. The models were trained using two GPUs with a total batch size of 512 and a maximum sequence length of 512.

\begin{figure}[t]
\centering
\includegraphics[width=\linewidth]{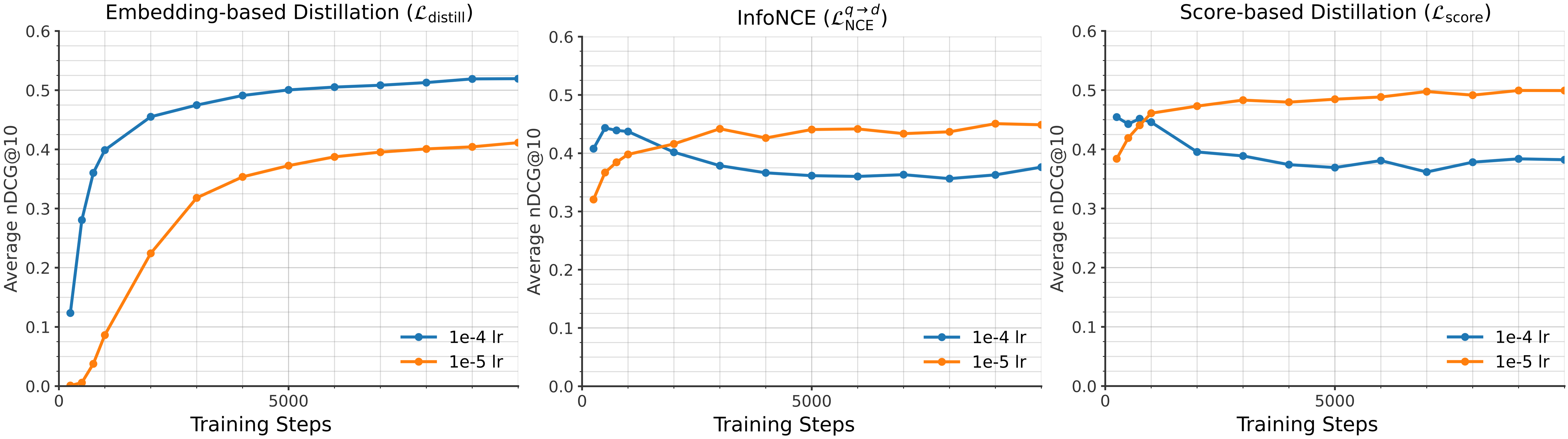}
\caption{\textbf{Learning rate sensitivity across different optimization objectives.} We report the average nDCG@10 on the MTEB (English, v2) benchmark using the \texttt{S2ORC} dataset. The plots compare $1\times10^{-4}$ (blue) and $1\times10^{-5}$ (orange) learning rates for embedding-based distillation ($\mathcal{L}_{\text{distill}}$), InfoNCE ($\mathcal{L}_{\text{NCE}}^{q \rightarrow d}$), and score-based distillation ($\mathcal{L}_{\text{score}}$), all utilizing a trainable student projection.}
\label{fig:lr-ablation}
\end{figure}

We investigated the sensitivity of InfoNCE ($\mathcal{L}_{\text{NCE}}$), feature-based distillation ($\mathcal{L}_{\text{distill}}$), and score-based distillation ($\mathcal{L}_{\text{score}}$) to two different learning rates: $1\times10^{-4}$ and $1\times10^{-5}$. The results, visualized in Figure~\ref{fig:lr-ablation}, reveal distinct optimization behaviors:

\begin{itemize}
    \item \textbf{Feature-based Distillation ($\mathcal{L}_{\text{distill}}$):} This objective is highly robust and performs significantly better with a higher learning rate of $1\times10^{-4}$. At $1\times10^{-5}$, convergence is considerably slower, and the model fails to reach the same performance ceiling within the same number of training steps.
    \item \textbf{InfoNCE ($\mathcal{L}_{\text{NCE}}^{q \rightarrow d}$):} In contrast, the contrastive objective is more sensitive to larger gradients. While a learning rate of $1\times10^{-4}$ shows a faster start, the performance eventually degrades or plateaus lower than the more stable $1\times10^{-5}$ run, which maintains better long-term consistency.
    \item \textbf{Score-based Distillation ($\mathcal{L}_{\text{score}}$):} Similar to InfoNCE, score-based distillation benefits from the lower learning rate. The $1\times10^{-4}$ configuration exhibits unstable behavior, with performance dropping sharply after an initial peak, whereas $1\times10^{-5}$ results in steady, sustained improvement.
\end{itemize}

Based on these observations, we selected $1\times10^{-4}$ for all $\mathcal{L}_{\text{distill}}$ experiments and $1\times10^{-5}$ for $\mathcal{L}_{\text{NCE}}$ and $\mathcal{L}_{\text{score}}$ in our main results to ensure each method is evaluated at its respective peak potential.

\newpage
%\begin{landscape}
%s
\subsection{Evaluation of MMTEB Performance Across Languages}
\label{app:language-eval}
\begin{figure}[H]
    \centering
    \includegraphics[width=0.85\linewidth]{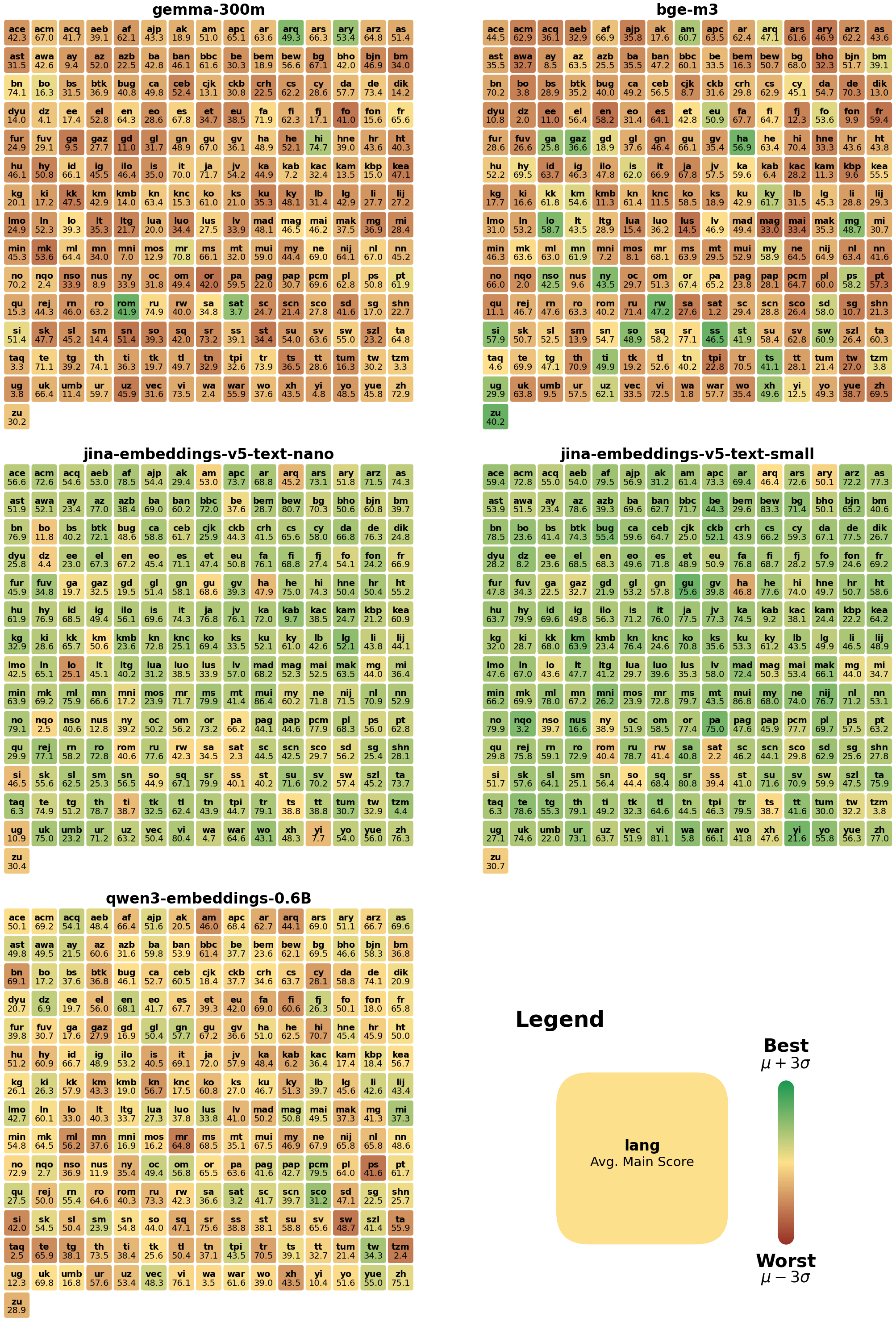}
    \caption{Performance of Models on different languages on MMTEB compared to average performance}
    \label{fig:langs}
\end{figure}
To visualize language-specific performance, we compute the mean and standard deviation of the per-language average scores on MMTEB. 
We map the interval $\mu \pm 3\sigma$ determined by the models performance for each language to a color scale. 
Figure~\ref{fig:langs} shows the resulting heatmaps.

%\end{landscape}
\newpage

\end{document}